\documentclass[letterpaper,11pt]{article}

\usepackage{lmodern}
\usepackage[T1]{fontenc}

\usepackage{geometry}
\usepackage{setspace}
\usepackage[articletitle = true, chaptertitle = true, doi = true]{achemso}

\usepackage[hyphens]{url}
\usepackage{graphicx}
\usepackage{float}
\newfloat{scheme}{htbp}{los}
\floatname{scheme}{Scheme}
\floatname{chart}{Chart}
\newfloat{graph}{htbp}{loh}

\usepackage[version = 4]{mhchem}   % formulas via \ce{}
\usepackage{amsmath}
\usepackage{amssymb}
\usepackage{amsthm}
\usepackage{algorithm}
\usepackage{algorithmic}
\usepackage{booktabs}
\usepackage{array}
\usepackage{caption}

\newtheorem{proposition}{Proposition}
\newtheorem{definition}{Definition}

\newcommand{\Spec}{\mathcal{S}}                 % spectrum
\newcommand{\pmz}{\rho}                          % precursor m/z
\newcommand{\Ad}{\mathcal{A}}                     % allowed adduct set
\newcommand{\f}{\mathbf{f}}                        % formula count vector
\newcommand{\fg}{\mathbf{f}^{\star}}              % correct formula
\newcommand{\nat}{\mathbb{N}}
\newcommand{\real}{\mathbb{R}}
\newcommand{\mass}{\mu}                            % neutral mass function
\newcommand{\mstar}{m^{*}}                         % neutral target mass
\newcommand{\rdbe}{\mathrm{rdbe}}
\newcommand{\Feas}{\mathcal{F}}                   % feasible set
\newcommand{\Enc}{\mathrm{Enc}_{\theta}}
\newcommand{\Hmat}{\mathbf{H}}                     % per-peak embeddings
\newcommand{\zvec}{\mathbf{z}}                     % pooled embedding
\newcommand{\score}{s_{\theta}}                    % scoring head

\usepackage{authblk}
\author{Xujun Che}
\author{Depeng Xu}
\author{Xiuxia Du*}
\affil{University of North Carolina at Charlotte, Charlotte, NC 28223, USA}

\title{FACET: Feasibility-Automaton Complete Enumeration and Truncation-Free Scoring of Molecular Formulas from MS/MS}
\date{*Email: xdu4@charlotte.edu}

\newtheorem{lemma}{Lemma}
\theoremstyle{plain}
\newtheorem*{propone}{Proposition 1 (Validity by construction)}
\newtheorem*{proptwo}{Proposition 2 (Recall guarantee)}

\usepackage{xcolor}
\definecolor{linkblue}{HTML}{1F4E8C}
\usepackage[colorlinks=true, linkcolor=linkblue, citecolor=linkblue, urlcolor=linkblue,
            pdftitle={FACET: Feasibility-Automaton Complete Enumeration and Truncation-Free Scoring of Molecular Formulas from MS/MS},
            pdfauthor={Xujun Che, Depeng Xu, Xiuxia Du}]{hyperref}

\begin{document}

\maketitle

\begin{abstract}
Determining the neutral molecular formula of a precursor ion from its tandem mass (MS/MS) spectrum is the first step in identifying an unknown small molecule. Existing methods trade recall, whether the correct formula is among the candidates, against the cost of examining each candidate. Methods that examine each candidate in depth must first shorten the list, by limiting element counts or truncating it, and that is where the correct formula is lost. End-to-end regressors are fast but guarantee neither validity nor recall. We present FACET, which separates deciding which formulas to consider from deciding which one to choose. A feasibility automaton builds a formula one element count at a time and admits exactly the formulas that fit the precursor mass and satisfy the ring-and-double-bond and valence constraints. Run as an exhaustive search, it enumerates the complete feasible set, so every candidate is valid and recall is fixed by the chemical constraints rather than by a trained model. A small neural scoring head then ranks every feasible formula from one pass of a pretrained spectrum encoder, so no candidate is discarded. On the NPLIB1 benchmark FACET reaches top-1 accuracies of $0.859$ with the adduct known and $0.729$ with it unknown, compared with $0.766$ and $0.660$ for the strongest established method, and its recall is the highest. On the external CASMI~2022 set its recall is $1.000$, and with the adduct unknown it is the most accurate of the methods compared.
\end{abstract}

\section{Introduction}

Untargeted metabolomics and natural-product discovery routinely produce millions of tandem mass spectra of unknown small molecules \citep{wang2016gnps,nothias2020fbmn,bushuiev2026dreams}, and only a small fraction of them can be annotated by matching against reference spectral libraries \citep{dasilva2015dark,horai2010massbank,che2026comparative}. For the remainder, identification has to begin from the spectrum itself, and the first question to answer is the molecule's neutral molecular formula, for example \ce{C20H30O4}, inferred from the mass-to-charge ratio ($m/z$) of the precursor ion and from the fragment peaks it produces \citep{kind2018review,blazenovic2018review}. An unequivocal formula is itself a reported level of identification confidence \citep{schymanski2014confidence}, and it is the starting point of almost every later step. Structure databases are queried with the formula \citep{duhrkop2015csi,ruttkies2016metfrag}, compound classes are predicted from fragmentation trees computed for it \citep{duhrkop2021canopus}, and de novo structure generators \citep{stravs2022msnovelist,che2026duallgd} are almost always given the formula and build the molecule on the atom set it fixes, so that a wrong formula becomes a wrong atom set \citep{che2026marlin}. The way a formula identifier fails therefore matters as much as how often it succeeds, and its failure modes are best excluded by construction.

Three properties make the problem hard. The first is \emph{validity}. A candidate formula is chemically possible only if its mass, corrected for the ion's adduct, matches the precursor within the instrument's mass accuracy; if its ring-and-double-bond equivalent (RDBE), the number of rings and double bonds the formula implies, is not negative \citep{pellegrin1983formulas}; and if its atoms carry enough valences to be joined into a single connected molecule \citep{senior1951partitions}. The adduct is the species added to or removed from the neutral molecule $M$ when it is ionized, such as a proton in $[M\!+\!\mathrm{H}]^{+}$, a sodium ion in $[M\!+\!\mathrm{Na}]^{+}$, or a proton together with the loss of a water molecule in $[M\!-\!\mathrm{H_2O}\!+\!\mathrm{H}]^{+}$. Plausibility criteria of this kind are standard filters in formula annotation \citep{kind2007seven}. The second property is \emph{recall}. Many formulas fit a given mass, even at sub-ppm accuracy \citep{kind2006mass}, and the correct one has to be among the candidates that are considered before any scoring can choose it. A precursor at $m/z$ $335.222$ observed as $[M\!+\!\mathrm{H}]^{+}$ and measured to within $30$\,ppm, for instance, admits $145$ distinct formulas inside its $\pm10$\,mDa window, and heavier precursors admit far more. Across the test spectra of the NPLIB1 benchmark, $72\%$ have more than $256$ feasible formulas, and above $800$\,Da the median feasible set holds about $360{,}000$. The third property is \emph{near-isobaric ambiguity}. The strongest competitors differ from the correct formula by an exchange of atoms of almost the same mass, so that mass alone cannot tell them apart and only the fragment spectrum can. In the example, \ce{C20H30O4} and \ce{C21H26N4} lie $1.34$\,mDa apart. In practice the most frequent of these exchanges are compound ones. Exchanging \ce{C3N} for \ce{O3H2} changes the mass by only $2.68$\,mDa, and exchanging \ce{C2OH2} for \ce{N3} by $1.34$\,mDa, whereas the textbook exchange of \ce{N} for \ce{CH2} changes it by $12.6$\,mDa. Some pairs are closer still: \ce{C3O} and \ce{NClH3} differ by $0.49$\,mDa.

Existing methods handle these properties in different ways, and they share a weakness that we call the \emph{recall--cost tension}: the more information a method extracts from the spectrum for each candidate, the more each candidate costs to score, and the fewer candidates the method can afford to consider. Combinatorial tools score candidates with expert-designed models of fragmentation, from heuristic rules combined with isotope-pattern matching \citep{pluskal2012formula} to hydrogen-rearrangement rules for annotating fragments \citep{tsugawa2016msfinder}. SIRIUS \citep{duhrkop2019sirius}, the most widely used of them, decomposes the precursor mass into every formula that fits, using the Round-Robin algorithm for the money-changing problem \citep{bocker2007money,duhrkop2013faster}. It ranks the candidates by isotope-pattern analysis \citep{bocker2009isotope} and by fragmentation trees \citep{bocker2008towards,rasche2011fragtrees,bocker2016reloaded}, which explain each fragment peak as a subformula of the candidate and must be computed separately for every candidate. ZODIAC then re-ranks the candidates jointly across all spectra of a data set \citep{ludwig2020zodiac}. BUDDY \citep{xing2023buddy} works bottom-up instead, assembling precursor formulas from the formulas that explain individual fragments and neutral losses and ranking them with machine-learned features and a false-discovery-rate estimate, and current versions of SIRIUS combine this bottom-up generation with decomposition. In all of these tools the candidate list is kept manageable by tight mass tolerances and by heuristic limits on element counts of the kind codified by Kind and Fiehn \citep{kind2007seven}.

Learned scoring reduces the hand-tuning but not the cost per candidate. MIST-CF \citep{goldman2024mistcf}, the most direct precedent for the present work, extends a spectrum Transformer \citep{goldman2023mist} into an energy-based ranker of (formula, adduct) pairs supplied by an external decomposer. Its authors list this reliance on the decomposition of SIRIUS among the limitations of the method. MIST-CF dispenses with the fragmentation tree but still assigns subformulas to the fragment peaks of every candidate, and this is expensive enough that the candidate list is shortened twice before scoring: the decomposer admits at most one atom of each halogen, and a fast pre-filter keeps only $256$ candidates. A formula removed at either step can never be ranked first, however good the scorer is. On NPLIB1, the halogen limit alone excludes the correct formula for $4.0\%$ of the test spectra. End-to-end regressors avoid candidate lists altogether. FIDDLE \citep{hong2025fiddle} predicts the formula directly with a network trained on about one million spectra and is fast and accurate at that scale, but nothing constrains its output to be chemically valid, and nothing ensures that the correct formula was ever considered.

\begin{table}[htbp]
\caption{How the compared methods generate and score candidate formulas. ``Valid'' means that every returned formula fits the precursor mass and satisfies an RDBE bound by construction.}
\label{tab:methods}
\centering
\small
\begin{tabular}{>{\raggedright\arraybackslash}p{1.6cm}>{\raggedright\arraybackslash}p{4.0cm}>{\raggedright\arraybackslash}p{4.35cm}>{\centering\arraybackslash}p{0.85cm}>{\centering\arraybackslash}p{1.9cm}}
\toprule
Method & Candidates & Scoring & Valid & Recall\newline guarantee \\
\midrule
SIRIUS~6 & mass decomposition below $m/z$ $400$ and bottom-up assembly from fragment formulas at every mass & a fragmentation tree for each candidate, of which the best $50$ are kept & yes & no \\
BUDDY & formulas retrieved by mass from a bundled database of $4.19$ million, and formulas assembled from fragment and neutral-loss formulas & machine-learned features of each candidate's fragment explanations & yes & no \\
MIST-CF & external decomposition by SIRIUS, at most one atom of each halogen & subformula-annotated peaks for the $256$ candidates kept by a learned formula filter & yes & no \\
FIDDLE & none, as the formula is regressed directly & refinement of the predicted formula & no & no \\
FACET & complete enumeration of the feasible set by the feasibility automaton & every feasible formula, scored by a small head on one shared spectrum encoding & yes & yes \\
\bottomrule
\end{tabular}
\end{table}

Table~\ref{tab:methods} summarizes where each approach obtains its candidates, what it scores, and what it guarantees. The method we present, FACET, removes the recall--cost tension by separating two decisions that the methods above couple: which formulas to consider, and which of them to choose. A \emph{feasibility automaton} makes the first decision. It builds a formula one element count at a time, discards at each step only counts that can no longer be completed into a valid formula, and accepts a finished formula only if it is valid. Because it never accepts an invalid formula and never discards a completable count, running it as an exhaustive search returns the complete set of valid formulas, and the correct formula belongs to that set whenever it is valid. FACET is database-independent: it considers every formula that the constraints admit, whether or not the formula occurs in a known compound. A neural network makes the second decision, and it is kept cheap per candidate: a spectrum encoder reads the spectrum once, and a scoring head about $450$ times smaller than the encoder ranks every formula in the set from that single encoding. Scoring the whole set, which for heavy precursors runs to millions of formulas, then becomes affordable, so FACET needs neither tight element limits nor truncation. The encoder is DreaMS \citep{bushuiev2026dreams}, which was pretrained on spectra alone and never sees a candidate formula, so the correct formula cannot leak into the spectrum representation.

Restricting what a generative model may output by means of a grammar or an automaton is an established technique in natural-language processing \citep{anderson2017constrained,scholak2021picard,willard2023outlines,geng2023grammar}, and in mass spectrometry a pruning rule on heavy-atom mass that provably never removes the true molecule already steers de novo structure generation toward the precursor mass \citep{che2026marlin}. The feasibility automaton applies the idea to formula identification with a constraint set that is exact, and it is this exactness that turns masking into a guarantee on recall. The contributions of this work are the following.

\begin{itemize}
\item \textbf{Validity by construction.} Every formula FACET emits satisfies the mass, RDBE and valence constraints without any filtering after the fact.
\item \textbf{A recall guarantee.} The same automaton enumerates the entire feasible set, so the correct formula is a candidate whenever it is feasible, however the candidates are later scored.
\item \textbf{Truncation-free scoring.} All candidates share one encoder pass, and the cost per candidate is that of a small scoring head, so FACET ranks the complete feasible set rather than a shortened list.
\item \textbf{A self-contained predictor.} FACET runs no external decomposition tool during training or inference, and even the negative examples it learns from come from its own enumeration.
\end{itemize}

On the NPLIB1 benchmark FACET ranks the correct formula first more often than any of four established methods when the adduct is known, and more often than the three that can run without the adduct when it has to be inferred. With the adduct known, its top-1 accuracy is $0.859$, compared with $0.766$ for the strongest of them. In both settings it also keeps the correct formula among its candidates more often. It keeps this lead for every adduct type and instrument represented by more than a handful of spectra, and its ranking advantage widens above $300$\,Da, where feasible sets become large. Its decomposer produces $160{,}000$ formulas per CPU second on a single thread, scales across processor cores, and runs inside the model. On the external CASMI~2022 challenge set the recall guarantee carries over unchanged.

\section{Theory}
\label{sec:theory}

\subsection{Problem Formulation}
\label{sec:problem}

A tandem mass spectrum $\Spec=\big(\pmz,\;\{(m_j,I_j)\}_{j=1}^{P}\big)$ consists of the precursor mass-to-charge ratio $\pmz$ and $P$ fragment peaks with $m/z$ values $m_j$ and intensities $I_j$. All spectra in this work are of singly charged positive ions, so an $m/z$ value is an ion mass. The ion is formed from the neutral molecule by an adduct $a\in\Ad$, and each adduct shifts the mass by a known amount $\delta_a$, which for $[M\!+\!\mathrm{H}]^{+}$ is the mass of a proton.

We fix an ordered set of elements $E=(e_1,\dots,e_K)$, with carbon first and hydrogen last. Element $e_k$ has monoisotopic mass $\nu_k$, valence $v_k$ and an upper bound $n_k$ on its count. A molecular formula is a vector of element counts $\f=(f_1,\dots,f_K)\in\nat^{K}$, and its neutral monoisotopic mass is
\begin{equation}
\mass(\f)=\textstyle\sum_{k=1}^{K} f_k\,\nu_k .
\label{eq:mass}
\end{equation}
Given $\pmz$ and $a$, the formula has to reproduce the neutral target mass $\mstar_a$ within the tolerance $\varepsilon_a$,
\begin{equation}
\mstar_a=\pmz-\delta_a,
\qquad
\varepsilon_a=\tau\,\mstar_a\,10^{-6},
\label{eq:target}
\end{equation}
where $\tau$ is the mass accuracy in parts per million (ppm). At $\tau=30$ and $\mstar_a=334$\,Da, for example, $\varepsilon_a$ is $10$\,mDa. The ring-and-double-bond equivalent counts the rings and double bonds a formula implies,
\begin{equation}
\rdbe(\f)=1+\tfrac{1}{2}\textstyle\sum_{k=1}^{K} f_k\,(v_k-2).
\label{eq:rdbe}
\end{equation}
For \ce{C20H30O4} it is $6$.

\begin{definition}[Feasible set]
\label{def:feasible}
For precursor $\pmz$ and adduct $a$, the \emph{feasible set} is
\begin{equation}
\Feas_a=\Big\{\,\f\in\nat^{K}\;:\;
|\mass(\f)-\mstar_a|\le\varepsilon_a,\;\;
\rdbe(\f)\ge 0,\;\;
\mathbf{0}\le\f\le\mathbf{n}
\Big\}.
\label{eq:feasible}
\end{equation}
\end{definition}

The RDBE condition is Senior's connectivity condition in another form. Senior's theorem \citep{senior1951partitions} requires that the valences of a molecule's atoms suffice to join them into one connected graph, $\sum_k f_k v_k \ge 2\left(\sum_k f_k-1\right)$, and rearranging eq~\ref{eq:rdbe} gives $\rdbe(\f)=\tfrac{1}{2}\big[\sum_k f_k v_k-2\left(\sum_k f_k-1\right)\big]$, so the two conditions coincide. Senior's second condition, that an even-electron molecule has an even total valence, is deliberately not imposed. Permanently charged cations, such as quaternary ammonium and pyridinium compounds like the phosphocholine \ce{C26H53NO7P}, are recorded with the formula of the cation, whose valence sum is odd and whose RDBE is a half-integer, so imposing parity would remove them from $\Feas_a$ and void the recall guarantee below. Parity is instead left to the scoring model as a soft signal. The feasible set that results, the mass window intersected with $\rdbe\ge0$ and the element bounds, is the filter that combinatorial decomposers apply, except that SIRIUS also admits an RDBE of $-0.5$. The recall comparisons in this work are therefore like for like.

We write $\fg$ for the correct formula. Candidate-generation \emph{recall} is the fraction of spectra whose correct formula is in the candidate set, and \emph{top-$k$ accuracy} is the fraction whose correct formula is among the $k$ highest-ranked candidates. The aim is a map $\Spec\mapsto\hat\f$, or $\Spec\mapsto(\hat a,\hat\f)$ when the adduct is unknown, that maximizes top-1 accuracy while guaranteeing that $\hat\f$ is valid.

\subsection{Overview of FACET}
\label{sec:method}

\begin{figure}[!t]
\centering
\includegraphics[width=\textwidth]{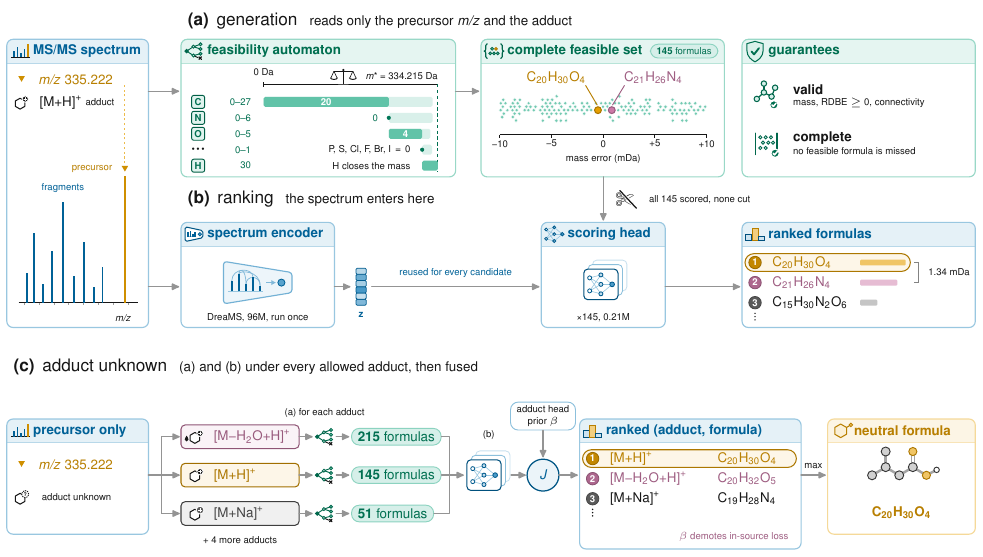}
\caption{FACET separates which formulas to consider from which to choose, illustrated for one precursor at $m/z$ $335.222$. (a)~Generation: the feasibility automaton admits element counts step by step and enumerates the $145$ formulas of the feasible set, shown as dots at their mass errors. (b)~Ranking: one encoder pass is shared by a small scoring head that scores every feasible formula. (c)~Adduct unknown: generation and ranking are repeated under every allowed adduct, and the (adduct, formula) pairs are ranked by a fused score.}
\label{fig:overview}
\end{figure}

FACET has two parts, illustrated in Figure~\ref{fig:overview} for the precursor at $m/z$ $335.222$. The first part generates candidates and carries the guarantees, and the second ranks them.

\emph{Generation}, shown in Figure~\ref{fig:overview}a, builds a formula one element count at a time in the fixed order $E$. At each step the feasibility automaton decides which counts are still admissible, discarding those that can no longer lead to a feasible formula. In the example, carbon may take any count from $0$ to $27$, because $28$ carbon atoms alone would exceed the target mass of $334.215$\,Da. Once $20$ carbon atoms are chosen, nitrogen may take $0$ to $6$ atoms, and after no nitrogen, oxygen may take $0$ to $5$. With $20$ carbon and $4$ oxygen atoms in place, the $30$\,Da that remain leave no room for phosphorus, sulfur, chlorine, bromine or iodine, whose masses all exceed it, but allow one fluorine atom. Hydrogen comes last, and its count is fixed by the mass that remains, here $30$ atoms. Running the automaton as an exhaustive search visits every admissible path and returns all $145$ feasible formulas. No trained model is involved in this part: generation reads only the precursor $m/z$, the adduct and the mass tolerance, and it never looks at the fragment peaks.

\emph{Ranking}, shown in Figure~\ref{fig:overview}b, is where the spectrum enters. A spectrum encoder reads the fragment peaks once. A small scoring head then assigns each of the $145$ formulas a score computed from that single encoding, and the highest-scoring formula is the prediction. In the example, \ce{C20H30O4} and \ce{C21H26N4} are only $1.34$\,mDa apart, so their order has to come from the fragments rather than from the mass.

The two parts decide different things: the automaton decides which formulas are considered, and the scoring head decides which of them wins. Recall does not depend on the scorer, and because the scorer is cheap it can afford to look at every candidate. When the adduct is unknown, both parts are repeated under every allowed adduct and the resulting (adduct, formula) pairs are ranked together, as Figure~\ref{fig:overview}c illustrates and the section on adduct-unknown inference below describes.

\subsection{The Feasibility Automaton}
\label{sec:automaton}

A formula is generated as the sequence of counts $f_1,f_2,\dots,f_K$ in the order $E$. After the first $k-1$ counts have been chosen, the state of the automaton is
\begin{equation}
\sigma_{k}=\Big(k,\;M_k=\!\!\textstyle\sum_{j<k} f_j\nu_j,\;\;V_k=\!\!\sum_{j<k} f_j(v_j-2)\Big),
\label{eq:state}
\end{equation}
that is, the mass accumulated so far and the accumulated contribution to the RDBE. Let $R_k=\sum_{j\ge k} n_j\nu_j$ be the largest mass that the elements from $e_k$ onward can still add. The mass test admits the count $f_k=c$ at a step $k<K$ if and only if
\begin{equation}
\big(\mstar_a-\varepsilon_a\big)-\big(M_k+c\,\nu_k\big)\le R_{k+1}
\;\;\wedge\;\;
M_k+c\,\nu_k\le \mstar_a+\varepsilon_a .
\label{eq:legal}
\end{equation}
In words, the running mass must not overshoot the upper end of the window, and it must not fall so far short of the lower end that even the heaviest possible completion could not reach it. Counts are also capped at $n_k$, and a second test, equally conservative, discards a count only when no completion could bring the RDBE back to zero or above. This test removes, for example, prefixes that already carry more halogen atoms than the chosen carbon, nitrogen and phosphorus atoms could bond.

Hydrogen, the last and lightest element, is closed deterministically: $f_K$ is the integer that brings $\mass(\f)$ closest to $\mstar_a$, and it is accepted only if the resulting formula lies within $\varepsilon_a$ and has $\rdbe(\f)\ge0$. Because the tolerance is far smaller than half the mass of a hydrogen atom, at most one hydrogen count can fall inside the window, so this closure loses nothing. The counts admitted at each step form a Boolean mask, which FACET also applies to the outputs of its autoregressive decoder, described below.

\begin{proposition}[Validity by construction]
\label{prop:validity}
Every complete formula accepted by the automaton lies in $\Feas_a$. Consequently every formula FACET outputs is chemically valid, without any filtering after the fact.
\end{proposition}
\begin{proof}[Proof sketch]
The tests of eq~\ref{eq:legal} and the count caps keep every prefix completable within $[\mstar_a-\varepsilon_a,\mstar_a+\varepsilon_a]$ and within $\mathbf{n}$, and the hydrogen closure enforces the mass window exactly and accepts a formula only when $\rdbe(\f)\ge0$. An accepted formula therefore satisfies every condition of eq~\ref{eq:feasible}.
\end{proof}

The tests at the non-final steps prune only prefixes that cannot be completed, and they never decide the exact RDBE, which the hydrogen closure settles. A feasible formula, the correct one included, is therefore never rejected along its own prefix, and this is what makes the enumeration of the next section lossless. Appendix~\ref{app:S2} proves this proposition and the next one in full, and Appendix~\ref{app:S3} traces the automaton through the example of Figure~\ref{fig:overview}.

\subsection{Exhaustive Enumeration and the Recall Guarantee}
\label{sec:enumeration}

Running the automaton as a depth-first search and keeping every accepted leaf enumerates the feasible set exactly, as Algorithm~\ref{alg:enum} sets out. FACET computes the same set with the Round-Robin algorithm for the money-changing problem \citep{bocker2007money}, the decomposition that SIRIUS also uses \citep{duhrkop2013faster}, implemented inside the model so that no external program is called. The algorithm scales the masses to integers and builds an extended residue table over the lightest element, hydrogen, which decides in constant time whether a partial sum can still be completed. Every solution is then checked against the exact masses and the RDBE condition. The two procedures return the same set.

\begin{algorithm}[t]
\caption{Feasible enumeration $\mathrm{Enum}(\pmz,a)$}
\label{alg:enum}
\textbf{Input}: precursor $\pmz$, adduct $a$ and element data $(\nu_k,v_k,n_k)$\\
\textbf{Output}: the feasible set $\Feas_a$
\begin{algorithmic}[1]
\STATE compute $\mstar_a,\varepsilon_a$ by eq~\ref{eq:target}; set $\Feas_a\leftarrow\varnothing$
\STATE \textbf{DFS}$(k{=}1,\;M{=}0,\;V{=}0,\;\f{=}\mathbf 0)$:
\IF{$k=K$ (hydrogen)}
  \STATE close $f_K$ deterministically; \textbf{if} $\f$ satisfies eq~\ref{eq:feasible} \textbf{then} add $\f$ to $\Feas_a$
\ELSE
  \FORALL{$c\le n_k$ admitted by eq~\ref{eq:legal} and by the RDBE test}
     \STATE \textbf{DFS}$(k{+}1,\,M{+}c\nu_k,\,V{+}c(v_k{-}2),\,\f\,[f_k{=}c])$
  \ENDFOR
\ENDIF
\STATE \textbf{return} $\Feas_a$
\end{algorithmic}
\end{algorithm}

\begin{proposition}[Recall guarantee]
\label{prop:recall}
$\mathrm{Enum}(\pmz,a)=\Feas_a$. Therefore, if $\fg\in\Feas_a$ then $\fg$ is in the candidate set, and candidate-generation recall equals $\Pr[\fg\in\Feas_a]$, independently of the scoring model and of any truncation.
\end{proposition}
\begin{proof}[Proof sketch]
Soundness is Proposition~\ref{prop:validity}. For completeness, neither the tests of eq~\ref{eq:legal} nor the conservative RDBE test ever prunes a prefix that can still be completed into a feasible formula, so every $\f\in\Feas_a$ is reached along its own prefix and accepted at its leaf.
\end{proof}

Proposition~\ref{prop:recall} makes recall a property of the chemical constraints rather than of the scoring model.

Feasible sets vary enormously in size, from a handful of formulas for light precursors to millions for heavy ones. FACET enumerates and scores them in full, at a cost that eq~\ref{eq:amortize} below makes precise. Enumeration runs under a fixed budget, and for the few spectra whose feasible set is estimated to exceed it, FACET falls back to constrained beam search \citep{anderson2017constrained,hokamp2017grid} under the same mask, which keeps validity but not completeness. Proposition~\ref{prop:recall} applies to every spectrum decoded on the exhaustive route.

\subsection{A Formula-Free Spectrum Encoder}
\label{sec:encoder}

FACET encodes the spectrum with DreaMS \citep{bushuiev2026dreams}, a Transformer \citep{vaswani2017attention} pretrained by self-supervision on millions of unannotated tandem mass spectra. DreaMS treats each of the $60$ most intense peaks, and the precursor, as a token described by its $m/z$ and intensity, and it returns an embedding for every peak and a pooled embedding of the whole spectrum,
\begin{equation}
(\Hmat,\zvec)=\Enc(\Spec),\qquad \Hmat\in\real^{P\times d},\;\;\zvec\in\real^{d}.
\label{eq:encoder}
\end{equation}
The encoder sees only the peak list and never a candidate formula, so the correct formula cannot leak into the representation. FACET fine-tunes the encoder end to end together with the rest of the model.

\subsection{Scoring the Complete Feasible Set}
\label{sec:scoring}

A small scoring head assigns each candidate a score computed from the pooled spectrum embedding, a learned embedding of the candidate's element counts, and two inexpensive features,
\begin{equation}
\score(\f\mid\Spec,a)=\mathrm{MLP}\!\Big(\zvec,\;\phi(\f),\;\tfrac{\Delta_{\mathrm{ppm}}(\f)}{10},\;\tfrac{\rdbe(\f)}{10}\Big),
\label{eq:score}
\end{equation}
where $\phi(\f)$ embeds the count vector and $\Delta_{\mathrm{ppm}}(\f)=10^{6}|\mass(\f)-\mstar_a|/\mstar_a$ is the candidate's mass error. The head is a three-layer perceptron with $0.21$ million parameters. Its mass features cannot separate near-isobars: two candidates that differ by an exchange of \ce{C2OH2} for \ce{N3} have almost the same mass error and similar RDBE. What separates them is the interaction between the spectrum embedding $\zvec$ and the composition embedding $\phi(\f)$, through which the head learns whether a composition is consistent with the fragmentation it observes. The prediction is
\begin{equation}
\hat\f=\arg\max_{\f\in\Feas_a}\;\score(\f\mid\Spec,a).
\label{eq:argmax}
\end{equation}

The score of eq~\ref{eq:score} is evaluated for every candidate, and no candidate is discarded. The cost of ranking splits into two parts,
\begin{equation}
\underbrace{\;\mathcal{O}(1)\;}_{\text{encoder, }96\text{M params}}\!\!\text{ per spectrum}
\;\;+\;\;
\underbrace{\;\mathcal{O}(|\Feas_a|)\;}_{\text{scorer, }0.21\text{M params}}\!\!\text{ over all candidates.}
\label{eq:amortize}
\end{equation}
The encoder, with $96$ million parameters, runs once per spectrum, however many candidates there are. The only cost per candidate is the scoring head, about $450$ times smaller, which assigns no subformulas to peaks. This is the structural difference from two-stage pipelines: they annotate the fragment peaks of each candidate separately and must prune or truncate the list, whereas FACET keeps every candidate at no extra encoder cost.

\subsection{The Autoregressive Decoder}
\label{sec:decoder}

FACET also contains an autoregressive decoder, a four-layer Transformer that predicts the element counts one at a time, in the order $E$, while attending to the per-peak embeddings $\Hmat$, with the automaton's mask applied at every step, so every formula it proposes is valid. The decoder extends FACET beyond the enumeration budget. For the rare spectra whose feasible set exceeds the budget, it supplies candidates by constrained beam search, and when the adduct is unknown, its log-likelihood also enters the fused ranking score. Because it shares the encoder pass with the scoring head, it adds little to the cost of training.

\subsection{Training Objective}
\label{sec:training}

With the adduct known, FACET is trained end to end on four terms,
\begin{equation}
\mathcal{L}=\mathcal{L}_{\mathrm{CE}}
+\lambda_{\mathrm{sem}}\,\mathcal{L}_{\mathrm{sem}}
+\lambda_{\mathrm{rank}}\,\mathcal{L}_{\mathrm{rank}}
+\lambda_{\mathrm{aux}}\,\mathcal{L}_{\mathrm{aux}}.
\label{eq:loss}
\end{equation}
$\mathcal{L}_{\mathrm{CE}}$ is the cross-entropy of the decoder's predicted counts under the automaton mask, computed with teacher forcing, in which the decoder sees the correct counts of the preceding elements at every step. Because the mask then depends only on the correct prefix, it is fixed for each training spectrum, and the unmasked outputs remain differentiable. $\mathcal{L}_{\mathrm{sem}}$ is minus the logarithm of the probability the decoder places on the admissible counts. It matters only early in training, while the mask is applied softly before it is annealed to a hard one. Teacher forcing is mixed with scheduled sampling \citep{bengio2015scheduled}, which increasingly replaces the correct prefix by the decoder's own choices, so that the decoder also learns to continue from its own mistakes. $\mathcal{L}_{\mathrm{aux}}$ regresses three coarse properties of the formula from the pooled embedding, namely the total number of atoms, the monoisotopic mass and the H/C ratio, which gives the encoder a dense training signal from the start.

The ranking loss $\mathcal{L}_{\mathrm{rank}}$ trains the scoring head. It is a softmax cross-entropy of the correct formula against a set of negatives \citep{oord2018cpc}, and the negatives are drawn from FACET's own enumeration of the training spectrum's feasible set, so they are near-isobaric formulas of the kind the head will face at inference. Half of them are the formulas closest in mass to the correct formula and half are sampled from the rest of the set, and no external decomposition tool takes part. When the adduct is unknown, a fifth term, the cross-entropy of the adduct head described below, is added with the same weight as $\mathcal{L}_{\mathrm{rank}}$. Checkpoints are selected by top-1 accuracy over the enumerated feasible sets of held-out validation spectra.

\subsection{Inference When the Adduct Is Unknown}
\label{sec:adductfree}

The adduct is often not known, for instance when MS1-level annotation \citep{kuhl2012camera} leaves a feature unresolved. FACET then infers it jointly with the formula, as Figure~\ref{fig:overview}c illustrates. An adduct head on the pooled embedding gives a posterior $q_{\theta}(a\mid\Spec)$ over the allowed adducts. FACET enumerates the feasible set under every allowed adduct, so the recall guarantee of Proposition~\ref{prop:recall} holds for the union of the sets, and it ranks (adduct, formula) pairs by the fused score
\begin{equation}
\begin{split}
J(a,\f)={}&\score(\f\mid\Spec,a)+w_{\mathrm{p}}\log q_{\theta}(a\mid\Spec)\\
          &+w_{\ell}\,\ell_\theta(\f\mid\Spec,a)+w_{\mathrm{e}}\,\beta(a),
\end{split}
\label{eq:af}
\end{equation}
where $\ell_\theta$ is the decoder's log-likelihood of the formula and $\beta(a)$ is a deterministic, parameter-free adduct-plausibility prior that equals $-1$ for the two in-source water-loss adducts, $[M\!-\!\mathrm{H_2O}\!+\!\mathrm{H}]^{+}$ and $[M\!-\!2\mathrm{H_2O}\!+\!\mathrm{H}]^{+}$, and $0$ for every other adduct. The prior expresses a preference for the simpler, intact reading of a spectrum. A water-loss adduct of one formula is the same ion as the protonated adduct of that formula minus \ce{H2O}, so the two readings always compete at exactly the same mass, and in-source loss is better treated as the exception than as the rule \citep{xu2015insource}. In the example of Figure~\ref{fig:overview}, the reading of the precursor as \ce{C20H32O5} with $[M\!-\!\mathrm{H_2O}\!+\!\mathrm{H}]^{+}$ has the same mass error as \ce{C20H30O4} with $[M\!+\!\mathrm{H}]^{+}$, and the prior demotes it. The three weights and a temperature for the adduct posterior are calibrated on the validation split, and the fused score is evaluated for the $80$ formulas that the scoring head ranks highest under each adduct. The neutral formula is finally obtained by maximizing over adducts, $\hat\f=\arg\max_{\f}\max_{a}J(a,\f)$.

\section{Experimental Section}
\label{sec:experiments}

\subsection{Data Sets}

We evaluate on NPLIB1 \citep{duhrkop2021canopus,goldman2024mistcf}, a benchmark of $10{,}709$ positive-mode MS/MS spectra annotated with their formula and adduct, covering $8{,}553$ distinct structures and $5{,}433$ distinct formulas. Seven adducts occur. The most common is $[M\!+\!\mathrm{H}]^{+}$, with $75\%$ of the spectra, followed by $[M\!+\!\mathrm{Na}]^{+}$, $[M\!+\!\mathrm{NH_4}]^{+}$, $[M\!-\!\mathrm{H_2O}\!+\!\mathrm{H}]^{+}$, $[M\!-\!2\mathrm{H_2O}\!+\!\mathrm{H}]^{+}$, $[M]^{+}$ and $[M\!+\!\mathrm{K}]^{+}$. Just over half of the spectra come from quadrupole time-of-flight (Q-ToF) instruments, $14\%$ from Orbitrap instruments and a few from ion traps, and a third carry no instrument annotation. Senior's parity condition would exclude the annotated formulas of $172$ of the $10{,}702$ spectra within FACET's element space, permanently charged cations recorded with the formula of the cation but annotated as $[M\!+\!\mathrm{H}]^{+}$ or another even-electron adduct. We use the standard split into $7{,}727$ training, $777$ validation and $2{,}205$ test spectra. The split partitions formulas rather than spectra, so no test formula and no test structure occurs in the training data, and every prediction on the test split concerns a formula the model has never seen. Eleven test spectra are excluded for every method: five whose formulas contain silicon or boron, outside the ten elements FACET decodes, and six whose peak lists the spectrum preprocessing could not convert into encoder input. All comparisons are paired over the remaining $2{,}194$ spectra.

For external validation we use CASMI~2022 \citep{casmi2022}, a round of the blind CASMI identification challenge \citep{schymanski2017casmi}. Its $304$ positive-mode spectra were all acquired on Orbitrap instruments, with $[M\!+\!\mathrm{H}]^{+}$, $[M\!+\!\mathrm{NH_4}]^{+}$ or $[M\!+\!\mathrm{Na}]^{+}$ adducts. We evaluate $295$ of them for every method. The other nine have precursors above $m/z$ $1{,}000$, beyond the input range of the spectrum encoder. CASMI~2022 differs markedly from the NPLIB1 training data, which are $53\%$ Q-ToF and only $13\%$ Orbitrap. For the experiment with additional training data we add $70{,}022$ Orbitrap spectra of $16{,}476$ structures from the NIST~23 tandem library \citep{nist2023} to the NPLIB1 training set. None of them carries a formula of the NPLIB1 test split, which is held out unchanged.

\subsection{Baselines}

We compare FACET with four methods that represent the approaches of Table~\ref{tab:methods}. All are evaluated on the same spectra against the same annotations. MIST-CF and FIDDLE are trained on the same NPLIB1 split as FACET, and SIRIUS and BUDDY are used as distributed. Appendix~\ref{app:S7} gives every setting.

SIRIUS~$6.5.4$ is run at its shipped settings, with the true adduct when the adduct is known and with the seven adducts of the data set when it is not. These settings generate candidates in two ways. A bottom-up search assembles formulas from fragment formulas at every precursor mass, without element restrictions, and below $m/z$ $400$ a de novo decomposition adds formulas over carbon, hydrogen, nitrogen, oxygen and phosphorus, with sulfur as a fallback when no isotope pattern is available. Other elements, such as chlorine or bromine, enter the decomposition only on isotope-pattern evidence, which these MS1-free spectra cannot provide. SIRIUS scores the candidates by fragmentation trees and reports the best $50$, so its recall is the recall of that list. MIST-CF, which scores up to $256$ SIRIUS decompositions per spectrum, is the stronger comparison built on SIRIUS.

The adduct-unknown setting is degenerate for SIRIUS. SIRIUS chooses the adduct from MS1 evidence, and without such evidence every adduct reading of one fragmentation tree receives the same score. For $77\%$ of the spectra, the best score is shared by several neutral formulas. SIRIUS breaks these ties by listing $[M\!+\!\mathrm{H}]^{+}$ first, which gives the right adduct for $65\%$ of the spectra, because three quarters of the split is $[M\!+\!\mathrm{H}]^{+}$.

BUDDY is run through \texttt{msbuddy}~$0.3.14$, its maintained implementation, with the model the package ships and with halogens enabled. It is the only machine-learned baseline not retrained here, so its scorer was trained on its authors' corpus rather than on our split. \texttt{msbuddy} does not decompose the precursor mass. It retrieves candidates by mass from a bundled database of $4.19$ million formulas, which contains $98.1\%$ of the correct formulas evaluated here, and adds the formulas it assembles from those that explain the fragments and neutral losses. BUDDY thus draws on knowledge of which formulas occur in known compounds, which the enumerating methods do not have and which, if anything, favors it. Its recall of $0.789$ is limited by the length of its own candidate list and lies well below that coverage.

MIST-CF runs its published pipeline. SIRIUS decomposes the precursor under the known adduct within $10$\,ppm and with its common element limits, which allow at most five sulfur, three phosphorus and one atom of each halogen. A learned filter that reads only the formula then keeps $256$ candidates, and the MIST-CF model scores them from their subformula-annotated fragment peaks. In the adduct-unknown setting SIRIUS decomposes the precursor under all seven adducts, and the filter keeps $256$ (formula, adduct) pairs across them.

FIDDLE takes the adduct and the collision energy as inputs, so it is evaluated with the adduct known only. NPLIB1 does not record collision energy, so a fixed $35$\,eV is supplied and the model runs without an input it was designed around. Its adduct scheme has no place for the $[M\!+\!\mathrm{K}]^{+}$ and $[M]^{+}$ spectra, which, together with ten spectra its encoder rejects, leaves $7{,}662$ of the $7{,}727$ training spectra. FIDDLE returns a formula for $2{,}154$ of the $2{,}194$ test spectra, so its low recall reflects the method rather than missing output: FIDDLE regresses one formula and refines it, instead of enumerating candidates.

Two inputs differ across methods and bear on how the numbers read. The first is the precursor value. For $1{,}410$ test spectra, almost two thirds, NPLIB1 records the precursor as the theoretical $m/z$ of the annotated ion, and for $750$ it records a measured value. For the remaining $34$, the recorded value is inconsistent with the annotated formula and adduct. FACET and FIDDLE read the measured value where one is recorded, SIRIUS and BUDDY read the parent mass of the spectrum file, and MIST-CF reads the precursor of its published benchmark, the theoretical $m/z$ with simulated noise of $\sigma=2.3$\,ppm. The second is the mass window. FACET enumerates within $30$\,ppm, and SIRIUS, BUDDY and MIST-CF within $10$\,ppm. On the precursor each method reads, the narrower window excludes seven correct formulas for SIRIUS and BUDDY, lowering their recall by $0.003$, and one for MIST-CF.

\subsection{Implementation}

FACET decodes ten elements in the order \ce{C}, \ce{N}, \ce{O}, \ce{P}, \ce{S}, \ce{Cl}, \ce{F}, \ce{Br}, \ce{I} and \ce{H}, with per-element maxima set from the training data. Silicon, boron and selenium are not decoded. The encoder is the self-supervised DreaMS model applied to the $60$ most intense peaks and fine-tuned end to end at one tenth of the decoder's learning rate. The decoder is a four-layer Transformer decoder with a model width of $512$, eight attention heads, a feed-forward width of $2{,}048$ and a dropout rate of $0.2$. The mass tolerance is $30$\,ppm for every spectrum, twice the largest instrument tolerance in the data, $15$\,ppm, and it is widened to $50$\,ppm only for an adduct under which no formula fits. Training uses Adam \citep{kingma2015adam} with a learning rate of $3\times10^{-4}$, a weight decay of $10^{-5}$ and batches of $16$ spectra, for at most $60$ epochs with early stopping. At inference the feasible set is enumerated exhaustively whenever its estimated size is at most $2\times10^{6}$ formulas, and constrained beam search of width $64$ serves as the fallback. Each model is trained and evaluated on a single NVIDIA L40S GPU. Appendix~\ref{app:S5} lists the remaining settings, including the loss weights, the mask schedule, the construction of the negatives and the calibrated fusion weights.

\subsection{Evaluation}

We report top-$k$ accuracy for $k=1$, $2$ and $3$, extended to $k=5$ and $10$ in Appendix~\ref{app:S8}, and candidate-generation recall. We also report the validity rate, the fraction of returned formulas that lie in the feasible set of the adduct used, and conditional top-1 accuracy, the top-1 accuracy over the spectra whose correct formula is among the candidates, which separates ranking quality from recall. A formula counts as correct only if every element count equals that of the annotated formula. Because all methods are evaluated on the same spectra, two methods are compared with McNemar's paired test \citep{mcnemar1947}, which considers only the spectra on which exactly one of them is correct, and accuracies by precursor mass are shown with Wilson score intervals \citep{wilson1927}.

\section{Results and Discussion}

\subsection{FACET Is More Accurate and More Complete on NPLIB1}

Table~\ref{tab:main} compares FACET with the four baselines on the $2{,}194$ test spectra. With the adduct known, FACET reaches a top-1 accuracy of $0.859$, and the strongest baseline, MIST-CF, reaches $0.766$. The comparison with MIST-CF can be made spectrum by spectrum: FACET is right where MIST-CF is wrong on $315$ spectra, and wrong where MIST-CF is right on $112$, a difference that McNemar's test places at $p=2\times10^{-23}$. The lead persists at deeper ranks, where FACET's top-3 accuracy is $0.914$ and that of MIST-CF $0.879$. FACET also has the highest recall of any method, $0.982$ compared with $0.959$ for MIST-CF, and every formula it returns is valid.

\begin{table}[htbp]
\caption{Top-$k$ accuracy and candidate-generation recall on the NPLIB1 test split ($n=2{,}194$).}
\label{tab:main}
\centering
\setlength{\tabcolsep}{5pt}
\begin{tabular}{llcccc}
\toprule
Adduct & Method & Top-1 & Top-2 & Top-3 & Recall \\
\midrule
Known   & SIRIUS         & 0.599 & 0.721 & 0.780 & 0.912 \\
        & FIDDLE         & 0.641 & 0.667 & 0.675 & 0.678 \\
        & BUDDY          & 0.736 & 0.767 & 0.776 & 0.789 \\
        & MIST-CF        & 0.766 & 0.847 & 0.879 & 0.959 \\
        & \textbf{FACET} & \textbf{0.859} & \textbf{0.896} & \textbf{0.914} & \textbf{0.982} \\
\midrule
Unknown & SIRIUS         & 0.448 & 0.540 & 0.574 & 0.905 \\
        & BUDDY          & 0.299 & 0.524 & 0.677 & 0.789 \\
        & MIST-CF        & 0.660 & 0.779 & 0.838 & 0.951 \\
        & \textbf{FACET} & \textbf{0.729} & \textbf{0.867} & \textbf{0.893} & \textbf{0.994} \\
\bottomrule
\end{tabular}
\end{table}

The recall column needs a word of interpretation, because the methods generate candidates so differently. SIRIUS's recall is that of its $50$ best candidates after fragmentation-tree scoring, BUDDY's that of its own candidate list, and FIDDLE's that of its refined predictions, which is why FIDDLE's top-3 accuracy and its recall nearly coincide. MIST-CF's recall is limited by its decomposer and its filter, and FACET's only by the few spectra, discussed below, whose recorded precursor places the correct formula outside the mass window or whose feasible set exceeds the enumeration budget.

With the adduct unknown, every method has to infer the adduct as well, and all accuracies drop. FACET still leads, with a top-1 accuracy of $0.729$ compared with $0.660$ for MIST-CF, a difference that McNemar's test again places far beyond chance, at $p=2\times10^{-11}$. Its top-2 accuracy of $0.867$ shows that the correct formula is usually its first or second choice. Its recall rises to $0.994$, higher than with the adduct known, because enumerating under every adduct also recovers spectra whose recorded precursor matches a different adduct than the annotated one, as discussed below. SIRIUS and BUDDY lose the most accuracy without the adduct, SIRIUS for the reason given in the Experimental Section.

\subsection{The Lead Holds on Measured Precursors}

NPLIB1 records the precursor of most test spectra as the theoretical $m/z$ of the annotated ion rather than as a measured value, and a theoretical precursor makes ranking unusually easy, because the correct formula then has a mass error of essentially zero. Ranking each enumerated feasible set by mass error alone puts the correct formula first for $1{,}397$ of the $1{,}403$ theoretical-precursor spectra whose correct formula is enumerated, but for only $127$ of the $748$ measured-precursor spectra. The measured-precursor spectra are therefore the fairest test of ranking quality, and on these $750$ spectra FACET still leads MIST-CF. With the adduct known, FACET reaches a top-1 accuracy of $0.771$ and MIST-CF $0.723$, a difference that is significant at $p=0.005$. With the adduct unknown, FACET leads by a similar margin, at $0.745$ compared with $0.711$ for MIST-CF. Appendix~\ref{app:S10} gives the accuracy of every method by precursor type.

\subsection{The Lead Holds Across Adducts, Instruments and Elements}

Table~\ref{tab:subsets} breaks the known-adduct accuracy down by adduct type, instrument and elemental composition. FACET's advantage over MIST-CF holds across the data. It holds for every adduct type with at least $35$ spectra, by margins from $0.066$ for $[M\!-\!\mathrm{H_2O}\!+\!\mathrm{H}]^{+}$ to $0.286$ for $[M\!-\!2\mathrm{H_2O}\!+\!\mathrm{H}]^{+}$, and for each instrument group in the table, including the third of the spectra that carry no instrument annotation. All of these differences are significant on a paired test, and Appendix~\ref{app:S8} gives the test statistics. Orbitrap spectra, which make up only $13\%$ of the training data, still show a wide margin, with FACET at $0.805$ and MIST-CF at $0.683$.

\begin{table}[htbp]
\caption{Top-1 accuracy on subsets of the NPLIB1 test split, adduct known. All methods are scored on the same spectra, and the best value in each row is in bold.}
\label{tab:subsets}
\centering
\small
\setlength{\tabcolsep}{5pt}
\begin{tabular}{lrccccc}
\toprule
Subset & $n$ & SIRIUS & FIDDLE & BUDDY & MIST-CF & FACET \\
\midrule
All spectra & $2{,}194$ & 0.599 & 0.641 & 0.736 & 0.766 & \textbf{0.859} \\
\addlinespace
$[M\!+\!\mathrm{H}]^{+}$ & $1{,}650$ & 0.630 & 0.614 & 0.702 & 0.751 & \textbf{0.833} \\
$[M\!+\!\mathrm{Na}]^{+}$ & $185$ & 0.427 & 0.670 & 0.751 & 0.768 & \textbf{0.897} \\
$[M\!+\!\mathrm{NH_4}]^{+}$ & $150$ & 0.393 & 0.873 & 0.847 & 0.833 & \textbf{0.973} \\
$[M\!-\!\mathrm{H_2O}\!+\!\mathrm{H}]^{+}$ & $167$ & 0.695 & 0.760 & 0.940 & 0.892 & \textbf{0.958} \\
$[M\!-\!2\mathrm{H_2O}\!+\!\mathrm{H}]^{+}$ & $35$ & 0.571 & 0.314 & 0.914 & 0.657 & \textbf{0.943} \\
\addlinespace
Bruker Q-ToF & $714$ & 0.633 & 0.856 & 0.838 & 0.882 & \textbf{0.966} \\
Q-ToF, other & $407$ & 0.553 & 0.528 & 0.828 & 0.715 & \textbf{0.853} \\
Orbitrap & $334$ & 0.656 & 0.464 & 0.623 & 0.683 & \textbf{0.805} \\
Not annotated & $718$ & 0.564 & 0.572 & 0.630 & 0.719 & \textbf{0.781} \\
\addlinespace
Containing sulfur & $250$ & 0.500 & 0.004 & 0.652 & 0.416 & \textbf{0.704} \\
\bottomrule
\end{tabular}
\end{table}

Elemental composition shows the recall guarantee at work. On the $87$ test spectra whose formula contains two or more atoms of one halogen, MIST-CF's decomposer cannot produce the correct formula at all, so its accuracy and recall are zero. FACET keeps every one of these formulas among its candidates and ranks $45\%$ of them first. Over all $221$ spectra with a halogenated formula, FACET's recall is $1.000$ and that of MIST-CF $0.606$. For formulas containing sulfur, FACET is the most accurate method, at $0.704$.

\subsection{Recall Holds and the Ranking Advantage Widens Above \texorpdfstring{300\,Da}{300 Da}}

\begin{figure}[htbp]
\centering
\includegraphics[width=0.63\textwidth]{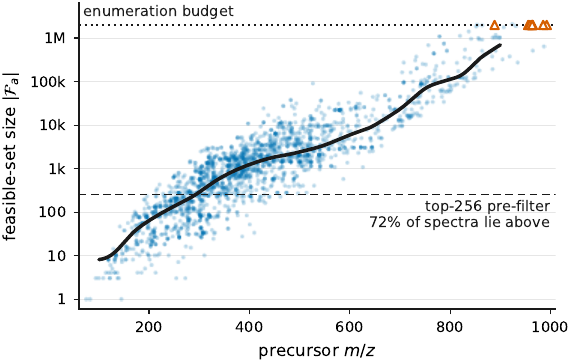}
\caption{Feasible sets grow by orders of magnitude with precursor mass. Size of the feasible set against precursor $m/z$, one point per NPLIB1 test spectrum, with the running median as a solid line. The dashed line marks $256$ candidates, the length of the list a pre-filter of that size would keep, and the dotted line marks the enumeration budget. Triangles mark the spectra whose sets exceed the budget.}
\label{fig:feasible}
\end{figure}

Feasible sets grow steeply with precursor mass. Figure~\ref{fig:feasible} plots their size for the NPLIB1 test spectra: the median set holds $22$ formulas below $200$\,Da and about $360{,}000$ above $800$\,Da, and it grows by a factor of $3.6$ for every additional $100$\,Da. With a budget of two million formulas, the exhaustive route handles $99.6\%$ of the test spectra when the adduct is known, and Proposition~\ref{prop:recall} applies to all of them. The other nine spectra all have precursors above $888$\,Da and go to beam search. Heavier precursors are therefore where truncation hurts most and where a ranker has the most candidates to tell apart.

Truncating these sets would cost recall that FACET keeps. Keeping only the $256$ feasible formulas closest in mass to the precursor would discard $4.6\%$ of the enumerated correct formulas, and $13.1\%$ of those whose precursor is a measured value. Even a list of $1{,}024$ candidates would lose $1.6\%$ and $4.7\%$, respectively. Limiting element counts, the other way to keep a list short, costs recall as well: it removes the correct formula of all $87$ multi-halogen spectra from the candidates of MIST-CF. FACET scores the complete feasible set and so keeps every enumerated correct formula.

\begin{figure}[htbp]
\centering
\includegraphics[width=0.9\textwidth]{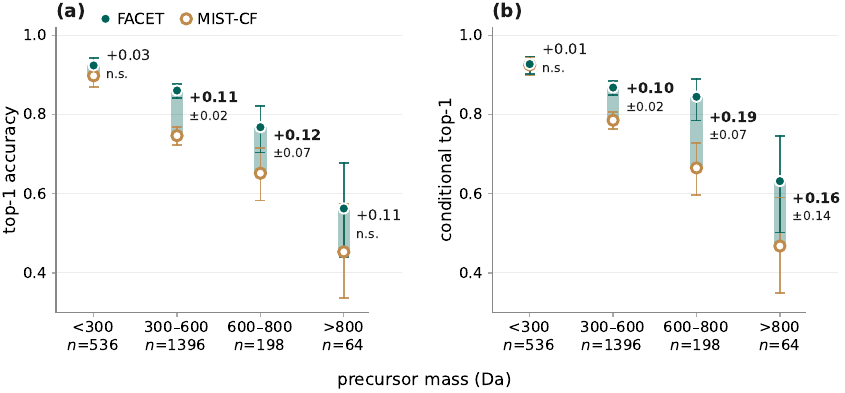}
\caption{FACET's ranking advantage widens above $300$\,Da. Accuracy of FACET and MIST-CF by precursor mass on NPLIB1, adduct known. (a)~Top-1 accuracy. (b)~Conditional top-1 accuracy, over the spectra whose correct formula is among the candidates. Points give each method's accuracy with its Wilson $95\%$ interval, and bars and labels give FACET's paired margin with its $95\%$ confidence interval for paired proportions (n.s., not significant).}
\label{fig:massbin}
\end{figure}

Figure~\ref{fig:massbin} compares FACET with MIST-CF in four mass ranges. Panel a shows top-1 accuracy. FACET is ahead in every range, and its margin is largest above $300$\,Da. Below $300$\,Da the margin is $0.03$, too small to be significant. It rises to $0.11$ from $300$ to $600$\,Da and to $0.12$ from $600$ to $800$\,Da, and both differences are significant on the paired test. Above $800$\,Da the margin is again $0.11$, but the $64$ spectra in that range are too few to establish it.

Panel b separates ranking from recall. It shows the conditional top-1 accuracy, which measures only how well each method ranks the candidates it has. FACET ranks better in every range, and here the margin is much wider above $300$\,Da, from $0.01$ below $300$\,Da to $0.10$, $0.19$ and $0.16$ in the three heavier ranges. The margin is significant everywhere except in the lightest range, where both methods are close to their ceiling.

The recall guarantee holds on every spectrum it covers. Every spectrum that FACET enumerates exhaustively and whose correct formula lies in its mass window has that formula in its candidate set, $2{,}151$ of $2{,}151$, as Proposition~\ref{prop:recall} requires, and across the whole of NPLIB1 the automaton admits at every step of its prefix the annotated formula of each of the $10{,}539$ spectra whose formula is feasible, as Appendix~\ref{app:S2} shows. On spectra with a consistent precursor, FACET's recall is therefore $1.000$ in every range below $800$\,Da, whereas that of MIST-CF lies between $0.951$ and $0.989$, and almost all of its misses are multi-halogen formulas that its decomposer excludes. FACET misses a correct formula only where a spectrum lies outside the guarantee: in five of the nine spectra above $888$\,Da whose feasible sets exceed the enumeration budget, and in the $34$ spectra whose recorded precursor is inconsistent with the annotated formula and adduct. The latter include $29$ permanently charged cations, $26$ of them phospholipids, annotated as $[M\!+\!\mathrm{H}]^{+}$ although their precursor is the mass of the $[M]^{+}$ cation, and the $18$ of them between $600$ and $800$\,Da lower FACET's recall in that range. MIST-CF recovers most of these spectra because its benchmark derives the precursor from the annotation, and FACET recovers $27$ of the $34$ when it also enumerates under $[M]^{+}$ with the adduct unknown, which is why its recall is higher without the adduct than with it. The comparison is thus decided by ranking, which carries FACET's top-1 accuracy above that of MIST-CF in every range.

\subsection{Most Remaining Errors Are Ones That Mass Cannot Resolve}

Most of the errors FACET still makes could not have been resolved by mass. We call a top-1 error mass-indistinguishable when the predicted and the correct formula both lie inside the same mass window, so that mass could not have separated them and only the spectrum could. With the adduct known, FACET ranks $310$ of the $2{,}194$ spectra wrongly, and $276$ of these errors are mass-indistinguishable, so what enumeration and ranking leave behind is almost entirely near-isobaric ambiguity. The exchanges behind these errors are compound and small. The most frequent, \ce{C3N} for \ce{O3H2}, accounts for a tenth of them and changes the mass by $2.68$\,mDa, and the next two, \ce{C2OH2} for \ce{N3} and \ce{CN4} for \ce{O4H4}, change it by only $1.34$\,mDa.

Of the $594$ errors FACET makes with the adduct unknown, $362$ assign the wrong adduct, and most of the others, $203$, confuse near-isobars under the correct adduct. Without the adduct, accuracy is thus limited mainly by the assignment of the adduct rather than by the ranking of formulas.

\subsection{Enumeration Is Fast and Scoring Stays Affordable}

Decomposition is where FACET and SIRIUS can be compared directly, because both implement the Round-Robin algorithm. Table~\ref{tab:eff} measures both on $120$ neutral target masses from the test split, with identical element bounds and a $30$\,ppm window, on one otherwise idle compute node. On a single thread, FACET's enumerator produces $160{,}000$ formulas per CPU second, and SIRIUS running on sixteen cores produces $125{,}000$. The two take about the same wall-clock time, $51$ and $48$\,s. Given the same sixteen cores, FACET finishes in $16$\,s, three times faster than SIRIUS, because its enumerator splits the mass window across threads, whereas the wall-clock time of SIRIUS does not respond to the number of cores it is given. FACET's enumerator also runs inside the model and passes its candidates straight to the scoring head, so decoding needs no external program. For the same window SIRIUS returns $11\%$ more formulas, mostly because it admits an RDBE of $-0.5$ where FACET requires zero or more. Appendix~\ref{app:S11} describes the benchmark in full.

\begin{table}[htbp]
\caption{Decomposition cost of FACET and SIRIUS~$6.5.4$ on $120$ neutral target masses from the NPLIB1 test split, with identical element bounds and a $30$\,ppm window.}
\label{tab:eff}
\centering
\setlength{\tabcolsep}{5pt}
\begin{tabular}{lccc}
\toprule
 & \multicolumn{2}{c}{FACET (in-process)} & SIRIUS~$6.5.4$ \\
\cmidrule(lr){2-3}\cmidrule(lr){4-4}
 & $1$ thread & $16$ threads & $16$ cores \\
\midrule
Formulas returned       & 8.2M & 8.2M & 9.1M \\
Formulas per CPU second & \textbf{160k} & 120k & 125k \\
Wall-clock time         & 51\,s & \textbf{16\,s} & 48\,s \\
\bottomrule
\end{tabular}
\end{table}

\begin{figure}[htbp]
\centering
\includegraphics[width=0.63\textwidth]{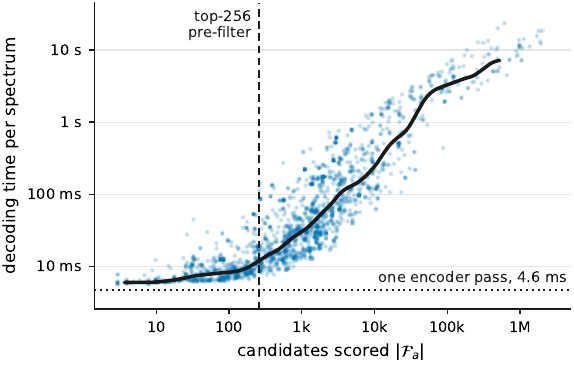}
\caption{Scoring every feasible formula stays affordable. FACET's decoding latency per spectrum against the number of candidates scored, NPLIB1 test split. Each point is one spectrum and the solid line is the running median. The dotted line is the cost of one encoder pass, and the dashed line marks $256$ candidates.}
\label{fig:latency}
\end{figure}

The cost of ranking grows more slowly than the number of candidates, because the encoder pass is shared. Figure~\ref{fig:latency} plots FACET's decoding time per spectrum against the number of candidates it scores. Below about $190$ candidates, scoring costs less than the single encoder pass of $4.6$\,ms, so a list of the size a top-$256$ pre-filter would keep costs FACET about as much as encoding the spectrum. Beyond that point the latency grows more slowly than the number of candidates, in proportion to $|\Feas_a|^{0.86}$. Scoring $10^5$ candidates takes a median of $3.4$\,s, and the largest feasible set of the split, $2.0$ million formulas, takes $18$\,s. A per-candidate subformula scorer has to cut any list longer than a few hundred candidates, and such lists are common: $72\%$ of the test spectra have more than $256$ feasible formulas.

\subsection{Ablations}

Table~\ref{tab:abl} isolates the components behind these results. The first block concerns the enumeration budget. Lowering it from two million to $200{,}000$ formulas, with the same trained model, routes more heavy spectra to beam search and lowers recall from $0.982$ to $0.969$ and top-1 accuracy from $0.859$ to $0.853$. The budget, not the scorer, therefore decides how many spectra the recall guarantee covers.

\begin{table}[htbp]
\caption{Ablations on the NPLIB1 test split.}
\label{tab:abl}
\centering
\setlength{\tabcolsep}{5pt}
\begin{tabular}{lcc}
\toprule
Variant & Top-1 & Recall \\
\midrule
\multicolumn{3}{l}{\emph{Enumeration budget, adduct known}} \\
FACET, $2\times10^{6}$ formulas & 0.859 & 0.982 \\
\quad $2\times10^{5}$ formulas & 0.853 & 0.969 \\
\addlinespace
\multicolumn{3}{l}{\emph{Fusion, adduct unknown}} \\
FACET, trained on NPLIB1 & 0.729 & 0.994 \\
\quad without the prior $\beta$ & 0.719 & 0.994 \\
FACET, trained on NIST~23 and NPLIB1 & 0.777 & 0.995 \\
\quad without the prior $\beta$ & 0.731 & 0.995 \\
\bottomrule
\end{tabular}
\end{table}

The second block concerns the fusion used when the adduct is unknown. The adduct-plausibility prior $\beta$ raises the top-1 accuracy of the model trained on NIST~23 and NPLIB1 from $0.731$ to $0.777$, and that of the model trained on NPLIB1 alone from $0.719$ to $0.729$, and both gains are significant on the paired test. Every spectrum that the prior corrects had been read as an in-source water-loss adduct without it. All variants emit only valid formulas, as Proposition~\ref{prop:validity} guarantees.

\subsection{Recall Transfers Unchanged to CASMI~2022}
\label{sec:external}

CASMI~2022 tests FACET under a change of instrument: its spectra are all from Orbitrap instruments, whereas the NPLIB1 training spectra are mostly Q-ToF, and every method reads the same measured precursor. Table~\ref{tab:casmi} reports top-1 accuracy in and out of domain. Its upper block compares the methods as trained on NPLIB1, with SIRIUS and BUDDY as distributed. Its lower block retrains FACET and MIST-CF on NPLIB1 together with the $70{,}022$ Orbitrap spectra from NIST~23, holding out the same NPLIB1 test split, so that the two are compared under matched training. FIDDLE is left out of the table. Its released code does not train the rescoring model that orders its refined candidates, so here they remain in order of mass error, and on measured precursors that order puts a near-isobar first.

\begin{table}[htbp]
\caption{Top-1 accuracy in domain on NPLIB1 and out of domain on CASMI~2022. Upper block: training on NPLIB1, with SIRIUS and BUDDY as distributed. Lower block: FACET and MIST-CF retrained on NIST~23 and NPLIB1. Best in each column in bold, second best underlined.}
\label{tab:casmi}
\centering
\begin{tabular}{ll cc cc}
\toprule
 & & \multicolumn{2}{c}{NPLIB1 (in domain)} & \multicolumn{2}{c}{CASMI 2022 (external)} \\
\cmidrule(lr){3-4}\cmidrule(lr){5-6}
Method & Training set & known & unknown & known & unknown \\
\midrule
SIRIUS  & distributed  & 0.599 & 0.448 & 0.780 & 0.620 \\
BUDDY   & distributed  & 0.736 & 0.299 & 0.790 & 0.414 \\
MIST-CF & NPLIB1      & 0.766 & 0.660 & 0.820 & 0.651 \\
FACET   & NPLIB1      & \underline{0.859} & \underline{0.729} & 0.814 & \underline{0.766} \\
\midrule
MIST-CF        & NIST+NPLIB1          & 0.832          & 0.681          & \textbf{0.881}     & 0.736 \\
FACET          & NIST+NPLIB1          & \textbf{0.889} & \textbf{0.777} & \underline{0.847}  & \textbf{0.810} \\
\bottomrule
\end{tabular}
\end{table}

The two halves of FACET respond to the shift in different ways, as the design predicts. The recall guarantee does not depend on training data at all: FACET's candidate-generation recall on CASMI is $1.000$ in both adduct settings and with either training set, the highest of any method. MIST-CF, whose decomposer and filter still discard correct formulas, reaches $0.976$ with the adduct known and $0.966$ without it, SIRIUS reaches $0.966$ in both settings, and FIDDLE, whose recall does not depend on how its candidates are ordered, reaches $0.766$. The learned ranker transfers as well, although it benefits from training data closer to the test domain. Trained on NPLIB1 alone, FACET reaches a CASMI top-1 accuracy of $0.814$ with the adduct known, close to the $0.820$ of MIST-CF, and $0.766$ with the adduct unknown, the best of any method in that block.

Under matched training on NIST~23 and NPLIB1, the comparison shows two things. First, FACET leads in the adduct-unknown setting both in and out of domain. Its top-1 accuracies are $0.777$ on NPLIB1 and $0.810$ on CASMI, compared with $0.681$ and $0.736$ for MIST-CF, so its accuracy with the adduct unknown does not degrade across the shift in instrument. Second, with the adduct known the ranking advantage depends on the domain. FACET leads in domain, where it reaches $0.889$ and MIST-CF $0.832$, whereas MIST-CF leads on CASMI, where it reaches $0.881$ and FACET $0.847$. With the adduct given, this comparison is pure near-isobar discrimination, and MIST-CF's explicit subformula assignment for each candidate transfers better to a new instrument than FACET's shared encoding, at the price of the correct formulas its filter removes.

\subsection{Limitations and Outlook}

FACET's advantage is carried by ranking and is largest where feasible sets are large, and so is its cost. Above $888$\,Da a feasible set can exceed the enumeration budget, and the nine spectra sent to beam search are the only ones whose candidate set falls short of the full feasible set. A larger budget, which the parallel enumerator makes affordable, would extend the guarantee to them. The guarantee is also bounded by the element space: it holds over the ten elements FACET decodes, so a molecule containing silicon, boron or selenium lies outside it. Extending the element space costs enumeration time but no change of method.

The external validation draws a clear line between the two halves of FACET. The recall guarantee, being a property of the constraints, transfers across instruments unchanged, whereas the learned ranker is sensitive to its training domain and improves with training data from the instrument of interest. More generally, the recipe of pairing an exhaustive candidate generator that carries a guarantee with a cheap scorer amortized over the whole candidate set applies to other identification problems with hard constraints.

\section{Conclusions}

Separating the decision of which formulas to consider from the decision of which to choose resolves the recall--cost tension that shapes current formula identification. Once candidate generation is exact and complete, recall is determined by the chemical constraints alone: it can be checked on every spectrum, and it carries over unchanged to new instruments. Once the expensive part of scoring is shared across candidates, ranking an entire feasible set of millions of formulas becomes affordable, and the correct formula is never removed before it is scored. On NPLIB1 the guarantees cost no accuracy: FACET is both more accurate and more complete than the methods that truncate. The errors that remain are near-isobaric exchanges that mass cannot resolve and, without the adduct, adduct misassignments, so further gains are to be found in reading the fragments more finely and in inferring the adduct. Because no feasible formula is lost, structure elucidation can start from a ranked list of formulas that contains the correct atom set whenever the chemistry admits it.

\section*{Acknowledgments}

The authors thank Dr.~Stephen Stein at the National Institute of Standards and Technology for providing the SDF version of the NIST~23 spectral library for this work. The work was made possible by UNC Charlotte internal funding.

% References
\bibliography{facet}

\clearpage
%% Supporting Information as lettered appendices; floats and lemmas are numbered within them.
%% The appendix is mostly tables: let them share pages with the text rather than stand on
%% half-empty float pages, and put any float page that remains at the top.
\renewcommand{\topfraction}{0.9}
\renewcommand{\bottomfraction}{0.8}
\renewcommand{\textfraction}{0.07}
\renewcommand{\floatpagefraction}{0.85}
\setcounter{topnumber}{3}
\setcounter{bottomnumber}{2}
\setcounter{totalnumber}{4}
\makeatletter
\setlength{\@fptop}{0pt}
\setlength{\@fpbot}{0pt plus 1fil}
\makeatother
\appendix
\counterwithin{table}{section}
\counterwithin{figure}{section}
\counterwithin{lemma}{section}
\pdfbookmark[0]{Appendix}{appendix}
\begin{center}
{\LARGE Appendix}
\end{center}
\bigskip
\noindent This appendix supports the main article. It explains the notation and the terms used there, proves the two propositions in full, follows the feasibility automaton through a worked example, and describes the enumeration, the model, its training, the data sets and the baseline configurations in more detail than the main text allows. It then reports the results behind the main text in full, including accuracy and recall for every method on subsets of the data, an analysis of the remaining errors and of how the precursor is recorded, the cost of truncating the candidate list, the computational benchmarks, and three representative test spectra.

Unless stated otherwise, all results are on the NPLIB1 test split. Of its $2{,}205$ spectra, $2{,}194$ are evaluated for every method, so every comparison between methods is paired. The eleven spectra left out are of two kinds. Five have annotated formulas that contain silicon or boron, outside the ten elements FACET decodes: three spectra of \ce{C16H15F2N3Si} and one each of \ce{C28H23BCl2F4N2O4} and \ce{C19H25BN4O4}. The other six have peak lists that the spectrum preprocessing could not convert into encoder input.

\section{Notation and Terminology}\label{app:S1}

Table~\ref{tab:s-notation} collects the symbols of the main text. The paragraphs after it explain the terms from mass spectrometry and from machine learning that the article relies on, for readers who are more familiar with one field than with the other.

\begin{table}[htbp]
\caption{Symbols used in the main text.}
\label{tab:s-notation}
\centering
\small
\begin{tabular}{>{\raggedright\arraybackslash}p{3.6cm}>{\raggedright\arraybackslash}p{11.4cm}}
\toprule
Symbol & Meaning \\
\midrule
$\mathcal{S}=(\rho,\{(m_j,I_j)\})$ & tandem mass spectrum: precursor $m/z$ $\rho$ and fragment peaks with $m/z$ $m_j$ and intensity $I_j$, $j=1,\dots,P$ \\
$a\in\mathcal{A}$, $\delta_a$ & adduct, from the set of allowed adducts, and the mass shift it introduces \\
$E=(e_1,\dots,e_K)$ & ordered element set, carbon first and hydrogen last \\
$\nu_k$, $v_k$, $n_k$ & monoisotopic mass, valence and maximum count of element $e_k$ \\
$\f=(f_1,\dots,f_K)$, $\fg$ & molecular formula as a vector of element counts; the correct formula \\
$\mu(\f)$ & neutral monoisotopic mass of $\f$ \\
$\mstar_a$, $\varepsilon_a$, $\tau$ & neutral target mass, mass tolerance in Da, mass accuracy in ppm \\
$\rdbe(\f)$ & ring-and-double-bond equivalent of $\f$ \\
$\Feas_a$ & feasible set: formulas in the mass window with $\rdbe\ge0$ and $\mathbf 0\le\f\le\mathbf n$ \\
$\sigma_k=(k,M_k,V_k)$ & automaton state after $k-1$ counts: step, accumulated mass, accumulated $\sum_{j<k}f_j(v_j-2)$ \\
$R_k$ & largest mass that elements $e_k,\dots,e_K$ can still add, $\sum_{j\ge k}n_j\nu_j$ \\
$\mathrm{Enc}_\theta$, $\mathbf H$, $\mathbf z$ & spectrum encoder, its per-peak embeddings and its pooled embedding \\
$s_\theta(\f\mid\mathcal S,a)$, $\phi(\f)$ & score of a candidate formula, and the learned embedding of its counts \\
$\Delta_{\mathrm{ppm}}(\f)$ & mass error of a candidate in ppm \\
$q_\theta(a\mid\mathcal S)$, $\ell_\theta$ & posterior of the adduct head, and decoder log-likelihood of a formula \\
$\beta(a)$ & adduct-plausibility prior, $-1$ for the two water-loss adducts and $0$ otherwise \\
$J(a,\f)$; $w_{\mathrm p}$, $w_\ell$, $w_{\mathrm e}$ & fused score used when the adduct is unknown, and its weights \\
\bottomrule
\end{tabular}
\end{table}

\paragraph{Precursor ion and tandem mass spectrum.} In tandem mass spectrometry (MS/MS) an ion of the intact molecule, the precursor, is selected by its mass-to-charge ratio ($m/z$) and fragmented, and the spectrum records the $m/z$ and the intensity of the fragment ions. MS1 denotes the spectrum of the intact ions before fragmentation, which carries the isotope pattern. The spectra used in this work contain MS/MS peaks only.

\paragraph{Adduct.} The precursor is not the neutral molecule $M$ itself but an ion formed from it. It can be the protonated molecule, $[M\!+\!\mathrm{H}]^{+}$, or the molecule with an attached sodium, ammonium or potassium ion, $[M\!+\!\mathrm{Na}]^{+}$, $[M\!+\!\mathrm{NH_4}]^{+}$ or $[M\!+\!\mathrm{K}]^{+}$. The protonated molecule can also lose one or two water molecules in the ion source, giving $[M\!-\!\mathrm{H_2O}\!+\!\mathrm{H}]^{+}$ or $[M\!-\!2\mathrm{H_2O}\!+\!\mathrm{H}]^{+}$. A permanently charged cation is recorded as $[M]^{+}$. The adduct fixes the mass difference $\delta_a$ between the neutral molecule and the precursor.

\paragraph{Mass accuracy and mass window.} Instruments measure $m/z$ with a relative error of a few parts per million (ppm). A tolerance of $\tau$\,ppm on a neutral mass $\mstar$ is a window of $\pm\tau\mstar\,10^{-6}$\,Da around it, and formulas whose monoisotopic mass falls in this window cannot be told apart by the precursor mass alone. The monoisotopic mass is the mass computed with the most abundant isotope of each element: \ce{^{12}C}, \ce{^{1}H}, \ce{^{14}N}, \ce{^{16}O}, \ce{^{31}P}, \ce{^{32}S}, \ce{^{35}Cl}, \ce{^{19}F}, \ce{^{79}Br} and \ce{^{127}I}.

\paragraph{Ring-and-double-bond equivalent and Senior's conditions.} The RDBE is the number of rings plus double bonds that a formula implies. For a formula \ce{C_cH_hN_nO_o} it equals $c-h/2+n/2+1$, with phosphorus counted like nitrogen and halogens like hydrogen. Senior's theorem gives conditions under which atoms of given valences can form a connected molecular graph. The connectivity condition, that the valence sum is at least twice the number of atoms minus two, is equivalent to $\rdbe\ge0$, as the main text shows. The parity condition, that the valence sum of an even-electron molecule is even, is not imposed by FACET, for the reason the main text gives.

\paragraph{Near-isobaric formulas.} Formulas whose masses differ by less than the mass window, typically through an exchange of atoms of almost equal total mass. Exchanging \ce{N} for \ce{CH2} changes the mass by $12.6$\,mDa, and exchanging \ce{C3O} for \ce{NClH3} by only $0.49$\,mDa.

\paragraph{Recall, top-$k$ accuracy and conditional top-1 accuracy.} Candidate-generation recall is the fraction of spectra whose correct formula is among the candidates a method generates. Top-$k$ accuracy is the fraction whose correct formula is among its $k$ highest-ranked candidates. Conditional top-1 accuracy is the top-1 accuracy over the spectra whose correct formula is among the candidates, so it measures ranking alone.

\paragraph{Encoder and Transformer.} An encoder is a neural network that maps an input, here a spectrum, to vectors that summarize it, called embeddings. A Transformer is a network that processes a set or a sequence of tokens, here the peaks of a spectrum, by letting every token attend to every other one.

\paragraph{Autoregressive decoding, teacher forcing and scheduled sampling.} An autoregressive decoder predicts a sequence one element at a time, each prediction conditioned on the previous ones. In FACET the sequence is the element counts of a formula. During training, teacher forcing gives the decoder the true previous counts, and scheduled sampling sometimes gives it its own earlier predictions instead, so that it also learns to continue from its own mistakes.

\paragraph{Beam search.} A search that keeps only the $B$ most probable partial sequences at every step. It is fast but can lose the correct sequence, which exhaustive enumeration cannot.

\paragraph{Contrastive ranking loss and hard negatives.} A ranking loss raises the score of the correct candidate relative to those of incorrect candidates, the negatives. Hard negatives are incorrect candidates that are difficult to distinguish from the correct one, and in FACET they are near-isobaric formulas from the spectrum's own feasible set.

\paragraph{McNemar's test and Wilson interval.} McNemar's test compares two methods evaluated on the same items by using only the items on which exactly one of them is correct. The Wilson score interval is a confidence interval for a proportion that stays accurate close to $0$ and $1$.

%% =====================================================================
\section{Proofs of Propositions 1 and 2}\label{app:S2}

We write out the automaton's tests in full. For a step $k<K$, the automaton admits the count $c$ at the state $\sigma_k=(k,M_k,V_k)$ if and only if
\begin{align*}
&\text{(i)}\;\; c\le n_k, \qquad
\text{(ii)}\;\; M_k+c\,\nu_k\le \mstar_a+\varepsilon_a, \qquad
\text{(iii)}\;\; (\mstar_a-\varepsilon_a)-(M_k+c\,\nu_k)\le R_{k+1},\\
&\text{(iv)}\;\; V_k+c\,(v_k-2)+U_{k+1}\ge-2, \quad\text{where } U_{k+1}=\textstyle\sum_{j>k} n_j\max(v_j-2,0).
\end{align*}
Tests (ii) and (iii) are the mass conditions of the main text, and test (iv) is the conservative connectivity test: $U_{k+1}$ is the largest increase of $\sum_j f_j(v_j-2)$ that the remaining elements could still contribute. At the last step, hydrogen, the automaton sets $f_K=\operatorname{round}\big((\mstar_a-M_K)/\nu_K\big)$ and accepts the formula if and only if $0\le f_K\le n_K$, $|M_K+f_K\nu_K-\mstar_a|\le\varepsilon_a$ and $V_K+f_K(v_K-2)\ge-2$. Throughout we use that $\rdbe(\f)\ge0$ is equivalent to $\sum_k f_k(v_k-2)\ge-2$, which follows directly from the definition of the RDBE.

\begin{lemma}[The hydrogen closure is unique]
\label{lem:h}
If $\varepsilon_a<\nu_{\mathrm H}/2$, at most one integer $h$ satisfies $|M+h\,\nu_{\mathrm H}-\mstar_a|\le\varepsilon_a$, and if one does, it equals $\operatorname{round}\big((\mstar_a-M)/\nu_{\mathrm H}\big)$.
\end{lemma}
\begin{proof}
Two integers that both satisfied the bound would give masses at least $\nu_{\mathrm H}>2\varepsilon_a$ apart inside a window of width $2\varepsilon_a$, which is impossible. If $h$ satisfies the bound, then $|(\mstar_a-M)/\nu_{\mathrm H}-h|\le\varepsilon_a/\nu_{\mathrm H}<1/2$, so $h$ is the nearest integer to $(\mstar_a-M)/\nu_{\mathrm H}$.
\end{proof}

The premise holds by a wide margin. With $\nu_{\mathrm H}=1.00783$\,Da, the lemma needs $\varepsilon_a<0.504$\,Da, whereas the largest window in this work, at $30$\,ppm and the heaviest test precursor of $m/z$ $992.5$, is $29.8$\,mDa. The implementation also keeps every window at least $2$\,mDa wide, a floor that binds only for targets below $67$\,Da and leaves the premise intact.

\begin{lemma}[The mass tests are necessary]
\label{lem:mass}
Let $\f\in\Feas_a$ and $k<K$. At the state reached along the prefix $(f_1,\dots,f_{k-1})$ of $\f$, the count $c=f_k$ passes tests (i) to (iii).
\end{lemma}
\begin{proof}
Test (i) holds because $\f\le\mathbf n$. All counts and masses are non-negative, so $M_k+f_k\nu_k=\sum_{j\le k}f_j\nu_j\le\mu(\f)\le\mstar_a+\varepsilon_a$, which is test (ii). Moreover $\mu(\f)-(M_k+f_k\nu_k)=\sum_{j>k}f_j\nu_j\le\sum_{j>k}n_j\nu_j=R_{k+1}$, and since $\mu(\f)\ge\mstar_a-\varepsilon_a$, test (iii) follows.
\end{proof}

\begin{lemma}[The connectivity test is necessary]
\label{lem:conn}
Let $\f\in\Feas_a$ and $k<K$. At the state reached along the prefix of $\f$, the count $c=f_k$ passes test (iv).
\end{lemma}
\begin{proof}
Because $\rdbe(\f)\ge0$, $\sum_j f_j(v_j-2)\ge-2$. The terms up to $k$ sum to $V_k+f_k(v_k-2)$, and the remaining terms satisfy $\sum_{j>k}f_j(v_j-2)\le\sum_{j>k}f_j\max(v_j-2,0)\le U_{k+1}$. Hence $V_k+f_k(v_k-2)+U_{k+1}\ge\sum_j f_j(v_j-2)\ge-2$.
\end{proof}

\begin{propone}
Every complete formula accepted by the automaton lies in $\Feas_a$.
\end{propone}
\begin{proof}
A formula is complete only after the hydrogen step. Its counts are non-negative by construction, $f_k\le n_k$ for $k<K$ by test (i), and $f_K\le n_K$ by the closure. The closure also requires $|\mu(\f)-\mstar_a|\le\varepsilon_a$ and $\sum_k f_k(v_k-2)\ge-2$, that is, $\rdbe(\f)\ge0$. Every condition of the definition of $\Feas_a$ therefore holds.
\end{proof}

\begin{proptwo}
$\mathrm{Enum}(\rho,a)=\Feas_a$. Consequently $\fg$ is a candidate whenever $\fg\in\Feas_a$, and candidate-generation recall equals $\Pr[\fg\in\Feas_a]$.
\end{proptwo}
\begin{proof}
Proposition~1 gives $\mathrm{Enum}(\rho,a)\subseteq\Feas_a$. For the converse, let $\f\in\Feas_a$. By Lemmas~\ref{lem:mass} and~\ref{lem:conn}, the count $f_k$ is admitted at every step $k<K$ of the depth-first search along the prefix of $\f$, so the search reaches the state formed by $(f_1,\dots,f_{K-1})$. There $|M_K+f_K\nu_K-\mstar_a|=|\mu(\f)-\mstar_a|\le\varepsilon_a$, so Lemma~\ref{lem:h} shows that the closure chooses exactly $f_K$, and it accepts $\f$ because $\f\le\mathbf n$ and $\rdbe(\f)\ge0$. Hence $\f\in\mathrm{Enum}(\rho,a)$. The statement about recall follows because the candidate set is $\mathrm{Enum}(\rho,a)$ on the exhaustive route.
\end{proof}

\paragraph{The Round-Robin computation.} FACET computes $\Feas_a$ with the Round-Robin decomposition rather than with the depth-first search of Algorithm~1. The decomposition visits every vector of non-negative counts within the bounds $\mathbf n$ whose integer-scaled mass lies in an interval that brackets the window, widened by the largest possible rounding error, and it uses an extended residue table over hydrogen to skip prefixes that cannot be completed. It then keeps exactly the vectors whose exact mass lies in the window and whose RDBE is non-negative, so it returns the same set, and Propositions~1 and~2 hold for it unchanged. The automaton's tests are used for the decoder's masks and for constrained beam search.

\paragraph{A direct check on NPLIB1.} We checked the completeness half of Proposition~2 directly on all NPLIB1 spectra: for each of the $10{,}539$ spectra whose correct formula lies in $\Feas_a$, the automaton admits the correct count at every step of its prefix, as Lemmas~\ref{lem:mass} and~\ref{lem:conn} require.

%% =====================================================================
\section{A Worked Example of the Feasibility Automaton}\label{app:S3}

Figure~1 of the main text follows one precursor through FACET. Here we trace the automaton through the same example with the numbers the implementation produces. The precursor has $m/z$ $335.2222$ and is observed as $[M\!+\!\mathrm{H}]^{+}$ with a $30$\,ppm tolerance, so the neutral target mass is $\mstar=334.2149$\,Da and the window extends $\pm10.03$\,mDa around it. Table~\ref{tab:s-trace} follows the automaton along the path of \ce{C20H30O4}.

\begin{table}[htbp]
\caption{The feasibility automaton along the path of \ce{C20H30O4} for a precursor at $m/z$ $335.2222$, $[M\!+\!\mathrm{H}]^{+}$, $30$\,ppm. ``Mass before'' is the mass of the counts already chosen, and ``room'' the mass that remains below the upper end of the window.}
\label{tab:s-trace}
\centering
\begin{tabular}{lcccc}
\toprule
Element & Mass before (Da) & Room (Da) & Admissible counts & Count taken \\
\midrule
\ce{C} & 0.000 & 334.225 & 0--27 & 20 \\
\ce{N} & 240.000 & 94.225 & 0--6 & 0 \\
\ce{O} & 240.000 & 94.225 & 0--5 & 4 \\
\ce{P} & 303.980 & 30.245 & 0 & 0 \\
\ce{S} & 303.980 & 30.245 & 0 & 0 \\
\ce{Cl} & 303.980 & 30.245 & 0 & 0 \\
\ce{F} & 303.980 & 30.245 & 0--1 & 0 \\
\ce{Br} & 303.980 & 30.245 & 0 & 0 \\
\ce{I} & 303.980 & 30.245 & 0 & 0 \\
\ce{H} & 303.980 & 30.245 & 30 & 30 \\
\bottomrule
\end{tabular}
\end{table}

Carbon comes first and may take any count from $0$ to $27$: $28$ carbon atoms alone would weigh $336$\,Da and overshoot the window, whereas for any smaller count the elements that follow could still add enough mass to reach the window, which is all that the mass test checks. Once $20$ carbon atoms are chosen, nitrogen may take $0$ to $6$ atoms, and with no nitrogen, oxygen may take $0$ to $5$. After $20$ carbon and $4$ oxygen atoms, $30.25$\,Da remain below the upper end of the window. Phosphorus, at $30.974$\,Da, and the heavier sulfur, chlorine, bromine and iodine no longer fit, while one fluorine atom, at $18.998$\,Da, still does. Hydrogen comes last and closes the mass with $30$ atoms. The exhaustive search over every admissible path returns $145$ formulas, all inside the window. \ce{C20H30O4} lies $-0.51$\,mDa from the target and \ce{C21H26N4} $+0.82$\,mDa, so the two are $1.34$\,mDa apart and only the spectrum can order them.

The same precursor gives different feasible sets under other adducts. Under $[M\!-\!\mathrm{H_2O}\!+\!\mathrm{H}]^{+}$ there are $215$ formulas, among them \ce{C20H32O5} at $-0.51$\,mDa: this is the in-source water-loss reading of the very same ion, with exactly the mass error of \ce{C20H30O4} under $[M\!+\!\mathrm{H}]^{+}$, and it is the reading that the prior $\beta$ demotes. Under $[M\!+\!\mathrm{Na}]^{+}$ there are $51$ formulas, among them \ce{C19H28N4} at $-1.58$\,mDa. When the adduct is unknown, FACET enumerates the feasible set under every allowed adduct in this way and ranks all the resulting (adduct, formula) pairs together.

%% =====================================================================
\section{Enumeration, Decoding Routes and the Enumeration Budget}\label{app:S4}

\paragraph{Enumeration.} FACET's enumerator implements the Round-Robin decomposition. Masses are scaled to integers by a factor of $100{,}000$, an extended residue table is built over hydrogen, the lightest element, and the search scans the integer masses that bracket the mass window, widened by the largest possible rounding error. Every solution is then checked against the exact masses and the RDBE condition, and the element bounds are enforced inside the recursion. The search is compiled to machine code and can run in parallel, with the integer masses of the window divided among threads. This is what lets FACET's enumerator use additional cores, as Appendix~\ref{app:S11} shows.

\paragraph{Decoding routes.} For every spectrum, and for every adduct when the adduct is unknown, FACET first counts the feasible set with an early exit at a budget of $2\times10^{6}$ formulas, and with a cap of $2\times10^{8}$ visited nodes. If the set fits the budget, it is enumerated exhaustively and scored in full. Otherwise FACET falls back to constrained beam search of width $64$, driven by the decoder under the automaton mask. If a route returns no candidate, FACET widens the beam to $256$, then relaxes the mass tolerance to $50$\,ppm and enumerates again, and only if that also fails does it return no answer for that adduct. FACET never returns a formula that violates the constraints, whatever the route, and it returns its $10$ highest-ranked formulas. Table~\ref{tab:s-routes} shows how often each route is taken and the recall achieved on it.

\begin{table}[htbp]
\caption{Share of test spectra decoded by each route and candidate-generation recall within the route ($n=2{,}194$). A dash marks a route that was never taken.}
\label{tab:s-routes}
\centering
\begin{tabular}{lcccc}
\toprule
 & \multicolumn{2}{c}{Adduct known} & \multicolumn{2}{c}{Adduct unknown} \\
\cmidrule(lr){2-3}\cmidrule(lr){4-5}
Route & Share (\%) & Recall & Share (\%) & Recall \\
\midrule
exhaustive enumeration & 99.59 & 0.984 & 98.36 & 0.997 \\
enumeration, relaxed tolerance & 0.00 & --- & 0.91 & 1.000 \\
constrained beam search & 0.41 & 0.444 & 0.73 & 0.625 \\
\bottomrule
\end{tabular}
\end{table}

With the adduct known, no spectrum needs the relaxed tolerance, and $99.6\%$ of the spectra take the exhaustive route, on which Proposition~2 applies. The remaining $9$ spectra all have precursors above $m/z$ $888$, and beam search recovers the correct formula for $4$ of them. With the adduct unknown, a spectrum is assigned to the beam route if any of its adducts needed beam search, and to the relaxed route if any needed the relaxed tolerance, so the shares are not directly comparable with those for the known adduct.

\paragraph{Feasible-set sizes.} Table~\ref{tab:s-feasible} summarizes the sizes of the feasible sets enumerated with the adduct known, which Figure~2 of the main text plots. The median grows by a factor of $3.6$ for every $100$\,Da of precursor mass, and $72\%$ of the enumerated sets hold more than $256$ formulas.

\begin{table}[htbp]
\caption{Sizes of the feasible sets enumerated with the adduct known, by precursor $m/z$.}
\label{tab:s-feasible}
\centering
\begin{tabular}{lrrrrc}
\toprule
Precursor $m/z$ & Spectra & Median & 90th percentile & Largest & Above $256$ (\%) \\
\midrule
$<200$ & 141 & 22 & 77 & 226 & 0 \\
200--400 & 1{,}072 & 364 & 1{,}642 & 10{,}513 & 58 \\
400--600 & 719 & 2{,}288 & 9{,}600 & 87{,}397 & 97 \\
600--800 & 198 & 24{,}870 & 163{,}076 & 1{,}042{,}603 & 100 \\
$>800$ & 55 & 362{,}367 & 1{,}552{,}008 & 1{,}997{,}469 & 100 \\
\bottomrule
\end{tabular}
\end{table}

\paragraph{The enumeration budget.} The enumeration budget, not the scorer, decides how many spectra the recall guarantee covers. With the same trained model, a budget of $2\times10^{5}$ formulas sends more heavy spectra to beam search and gives a recall of $0.969$ and a top-1 accuracy of $0.853$, whereas the budget of $2\times10^{6}$ used throughout gives $0.982$ and $0.859$.

%% =====================================================================
\section{Model Architecture and Training}\label{app:S5}

\paragraph{Architecture.} Table~\ref{tab:s-arch} lists the modules of FACET and their parameter counts, read from the trained checkpoint. The encoder is DreaMS, pretrained by self-supervision and fine-tuned here. It reads the $60$ most intense peaks of a spectrum together with the precursor, and the output at the precursor token serves as the pooled embedding $\mathbf z$. The decoder is a Transformer decoder with $4$ layers, a model width of $512$, $8$ attention heads, a feed-forward width of $2048$ and a dropout rate of $0.2$. Its input at step $k$ is the sum of an embedding of the step, an embedding of the count chosen at step $k-1$, and a projection of embeddings of the adduct and the instrument type. It attends causally to the previous steps and, through cross-attention, to the per-peak embeddings $\mathbf H$. A linear layer shared by all steps maps the decoder output to logits over the possible counts, and the automaton mask restricts them to the admissible ones. The scoring head maps the count vector, scaled by $1/10$, through a linear layer to a $64$-dimensional embedding $\phi(\f)$, concatenates it with $\mathbf z$, the scaled mass error and the scaled RDBE, and applies a three-layer perceptron with GELU activations and $256$ hidden units. The auxiliary head regresses the total number of atoms, the monoisotopic mass and the H/C ratio from $\mathbf z$, and the adduct head predicts the adduct from $\mathbf z$.

\begin{table}[htbp]
\caption{Modules of FACET and their parameter counts.}
\label{tab:s-arch}
\centering
\begin{tabular}{llr}
\toprule
Module & Role & Parameters (M) \\
\midrule
DreaMS encoder & spectrum embeddings $\mathbf H$, $\mathbf z$; once per spectrum & $96.080$ \\
Autoregressive decoder & beam-search fallback, log-likelihood $\ell_\theta$ & $16.955$ \\
Scoring head & score of every candidate & $0.215$ \\
Auxiliary head & atom count, mass and H/C regression & $0.132$ \\
Adduct head & adduct posterior $q_\theta$ & $0.133$ \\
\midrule
Total & & $113.5$ \\
\bottomrule
\end{tabular}
\end{table}

The encoder has about $447$ times as many parameters as the scoring head, which is the only module whose cost grows with the number of candidates.

\paragraph{Element space.} FACET decodes ten elements in the order and with the maximum counts \ce{C}~$74$, \ce{N}~$15$, \ce{O}~$33$, \ce{P}~$6$, \ce{S}~$9$, \ce{Cl}~$7$, \ce{F}~$21$, \ce{Br}~$4$, \ce{I}~$6$, \ce{H}~$128$. The maxima were set from the training data with headroom, and Appendix~\ref{app:S6} compares them with the largest counts observed in NPLIB1. Silicon, boron and selenium are not decoded.

\paragraph{Training.} Table~\ref{tab:s-train} lists the training settings of the known-adduct model. The mass tolerance during training and inference is the instrument tolerance raised to a floor of $30$\,ppm, which is $30$\,ppm for every spectrum, apart from the relaxed route of Appendix~\ref{app:S4}, because no instrument tolerance in the data set exceeds $15$\,ppm.

\begin{table}[htbp]
\caption{Training settings of the known-adduct model.}
\label{tab:s-train}
\centering
\begin{tabular}{l>{\raggedright\arraybackslash}p{10.4cm}}
\toprule
Setting & Value \\
\midrule
Optimizer & Adam, learning rate $3\times10^{-4}$, weight decay $10^{-5}$ \\
Encoder learning rate & $0.1\times$ the decoder learning rate \\
Batch size; epochs & $16$ spectra; at most $60$, early stopping after $12$ epochs without improvement \\
Loss weights & $\lambda_{\mathrm{sem}}=0.1$, $\lambda_{\mathrm{rank}}=0.3$, $\lambda_{\mathrm{aux}}=0.3$ \\
Label smoothing & $0.05$ on the decoder cross-entropy, $0.05$ on the ranking loss \\
Automaton mask & soft for $2$ epochs, annealed to hard over the next $2$ \\
Scheduled sampling & from epoch $4$, probability ramped up to $0.25$ \\
Negatives per spectrum & $64$ per training step, from a pool of $96$ \\
Checkpoint selection & top-1 accuracy over enumerated validation feasible sets \\
\bottomrule
\end{tabular}
\end{table}

\paragraph{Hard negatives.} For each training and validation spectrum, FACET enumerates the feasible set under the true adduct and keeps a pool of $96$ negatives: the $48$ formulas closest in mass to the correct formula and $48$ sampled at random from the rest. The pool is restricted to formulas with at most five sulfur, three phosphorus and one atom of each halogen, because unrestricted near-isobaric enumerations are dominated by halogen-rich formulas that make the contrast too easy. Pools exist for $8{,}058$ of the $8{,}504$ training and validation spectra. The others have feasible sets too large to enumerate within the budget used for building the pools. The restriction applies only to the training negatives: the candidate sets ranked at inference are never restricted.

\paragraph{The adduct-unknown fusion.} The temperature of the adduct posterior and the weights $w_{\mathrm p}$, $w_\ell$ and $w_{\mathrm e}$ of the fused score are chosen by a grid search that maximizes top-1 accuracy on the validation split. For the model trained on NPLIB1 the search selects a temperature of $2$, $w_{\mathrm p}=1$, $w_\ell=0.1$ and $w_{\mathrm e}=0.5$, which give a validation top-1 accuracy of $0.783$. For the model trained on NIST~23 and NPLIB1 the selected values are $0.75$, $0.5$, $0.1$ and $1$, with a validation top-1 accuracy of $0.813$. Under each adduct, the fused score is evaluated for the $80$ formulas that the scoring head ranks highest, and recall is counted over the full union of the feasible sets.

\paragraph{Hardware.} Each model is trained and evaluated on a single NVIDIA L40S GPU with $8$ CPU cores and $50$\,GB of memory.

%% =====================================================================
\section{Data Sets}\label{app:S6}

\paragraph{NPLIB1.} NPLIB1 holds $10{,}709$ positive-mode MS/MS spectra of $8{,}553$ distinct structures and $5{,}433$ distinct formulas. Tables~\ref{tab:s-adducts} and~\ref{tab:s-instr} give the adducts and the instrument types in each part of the standard split. The split partitions formulas: the training, validation and test parts contain $3{,}912$, $434$ and $1{,}087$ distinct formulas, and the test part shares none of them with the others, so every test formula is new to the model.

\begin{table}[htbp]
\caption{Adducts in the NPLIB1 split.}
\label{tab:s-adducts}
\centering
\begin{tabular}{lrrrr}
\toprule
Adduct & Training & Validation & Test & Total \\
\midrule
$[M\!+\!\mathrm{H}]^{+}$ & 5{,}785 & 588 & 1{,}657 & 8{,}030 \\
$[M\!+\!\mathrm{Na}]^{+}$ & 741 & 69 & 188 & 998 \\
$[M\!+\!\mathrm{NH_4}]^{+}$ & 507 & 61 & 150 & 718 \\
$[M\!-\!\mathrm{H_2O}\!+\!\mathrm{H}]^{+}$ & 497 & 43 & 167 & 707 \\
$[M\!-\!2\mathrm{H_2O}\!+\!\mathrm{H}]^{+}$ & 142 & 12 & 35 & 189 \\
$[M]^{+}$ & 41 & 1 & 5 & 47 \\
$[M\!+\!\mathrm{K}]^{+}$ & 14 & 3 & 3 & 20 \\
\midrule
All & 7{,}727 & 777 & 2{,}205 & 10{,}709 \\
\bottomrule
\end{tabular}
\end{table}

\begin{table}[htbp]
\caption{Instrument types in the NPLIB1 split, as annotated in the data set.}
\label{tab:s-instr}
\centering
\begin{tabular}{lrrrr}
\toprule
Instrument & Training & Validation & Test & Total \\
\midrule
Bruker Q-ToF & 2{,}478 & 264 & 716 & 3{,}458 \\
Q-ToF, other & 1{,}598 & 136 & 410 & 2{,}144 \\
Orbitrap & 1{,}021 & 102 & 338 & 1{,}461 \\
ion trap & 61 & 9 & 21 & 91 \\
FT-ICR & 1 & 0 & 0 & 1 \\
not annotated & 2{,}568 & 266 & 720 & 3{,}554 \\
\bottomrule
\end{tabular}
\end{table}

Table~\ref{tab:s-elements} shows how often each element occurs and the largest count observed, next to the maximum count FACET admits. Every NPLIB1 formula within the element space fits the bounds. Carbon reaches $66$ atoms, below its bound of $74$, and hydrogen $112$, below its bound of $128$.

\begin{table}[htbp]
\caption{Elements in the NPLIB1 formulas: share of formulas that contain the element, largest count observed, and the maximum count FACET admits.}
\label{tab:s-elements}
\centering
\begin{tabular}{lccc}
\toprule
Element & Share of formulas (\%) & Largest count & FACET bound \\
\midrule
\ce{N} & 51.9 & 11 & 15 \\
\ce{O} & 96.0 & 27 & 33 \\
\ce{P} & 3.0 & 3 & 6 \\
\ce{S} & 9.0 & 6 & 9 \\
\ce{Cl} & 6.6 & 4 & 7 \\
\ce{F} & 3.2 & 17 & 21 \\
\ce{Br} & 0.5 & 2 & 4 \\
\ce{I} & 0.3 & 4 & 6 \\
\bottomrule
\end{tabular}
\end{table}

\paragraph{CASMI~2022.} The positive-mode set has $304$ spectra, all acquired on Orbitrap instruments, with the adducts $[M\!+\!\mathrm{H}]^{+}$ 177, $[M\!+\!\mathrm{NH_4}]^{+}$ 67, $[M\!+\!\mathrm{Na}]^{+}$ 60. We evaluate $295$ of them for every method. The other $9$, with correct formulas \ce{C35H44I6N6O15}, \ce{C48H86O25}, \ce{C49H88O25}, \ce{C53H84O23}, \ce{C54H80O27}, \ce{C54H84O22}, \ce{C54H90N6O18}, \ce{C58H104O28}, have precursors of $m/z$ $1{,}081$ and above, beyond the input range of the spectrum encoder, which accepts precursors up to $m/z$ $1{,}000$.

\paragraph{NIST~23.} The additional training data are $70{,}022$ Orbitrap spectra of $16{,}476$ distinct structures from the NIST~23 tandem library, with the adducts $[M\!+\!\mathrm{H}]^{+}$ 45{,}006, $[M\!-\!\mathrm{H_2O}\!+\!\mathrm{H}]^{+}$ 11{,}987, $[M\!+\!\mathrm{Na}]^{+}$ 9{,}354, $[M\!-\!2\mathrm{H_2O}\!+\!\mathrm{H}]^{+}$ 1{,}976, $[M\!+\!\mathrm{NH_4}]^{+}$ 1{,}402, $[M\!+\!\mathrm{K}]^{+}$ 297. None of them carries a formula of the NPLIB1 test split, which is held out unchanged, so the in-domain comparison of models trained with NIST~23 remains a comparison on unseen formulas.

%% =====================================================================
\section{Baseline Configurations}\label{app:S7}

\paragraph{SIRIUS.} SIRIUS~$6.5.4$ is run through its command-line interface, with \texttt{--ppm-max=10}, \texttt{-c=50} to keep $50$ candidates per spectrum, and a timeout of $600$\,s per compound. With the adduct known, the considered adducts are pinned to the annotated one, and with the adduct unknown they are the seven adducts of the data set. All other settings are the shipped defaults. Under these defaults, candidates come from a bottom-up search at every precursor mass, which is not restricted by the element alphabet, and from a de novo decomposition below $m/z$ $400$. The decomposition always considers C, H, N, O and P, adds sulfur as a fallback when no isotope pattern is available, and considers S, Br, Cl, B, Se, Fe, Zn and Mg only when an isotope pattern indicates them, which cannot happen for these MS1-free spectra. SIRIUS's de novo decomposition admits an RDBE of $-0.5$, one half below FACET's bound. When the adduct is unknown and several adducts tie on the same fragmentation tree, SIRIUS lists $[M\!+\!\mathrm{H}]^{+}$ first, as the main text describes.

\paragraph{BUDDY.} BUDDY is run through \texttt{msbuddy}~$0.3.14$ with the model the package ships, a tolerance of $10$\,ppm for both the precursor and the fragments, and halogens enabled. Each spectrum's peak list is reduced to its $256$ most intense peaks. With the adduct known, every spectrum is submitted with its annotated adduct. With the adduct unknown, it is submitted once under each of the data set's adducts, and the candidates from all adducts are ranked together by their scores. \texttt{msbuddy} retrieves candidates by mass from a bundled database of formulas of known compounds, adds the formulas it assembles from those that explain the fragments and neutral losses, and keeps at most $350$ of them for each adduct it is run under.

\paragraph{MIST-CF.} MIST-CF runs its published pipeline with a model trained on the NPLIB1 training split. SIRIUS decomposes the precursor within $10$\,ppm, under the annotated adduct when it is known and under all seven adducts when it is not, with the common element limits of at most five sulfur, three phosphorus and one atom of each of fluorine, chlorine, bromine and iodine. A learned filter that reads only the formula keeps $256$ candidates, subformulas are assigned to the $100$ most intense fragment peaks of each candidate, and the MIST-CF model scores the candidates. MIST-CF reads the precursor of its published benchmark, the theoretical $m/z$ of the annotated ion with simulated noise.

\paragraph{FIDDLE.} FIDDLE is retrained on the NPLIB1 training split with its own preprocessing, which keeps $7{,}662$ of the $7{,}727$ training spectra: its adduct scheme has no place for the $55$ $[M\!+\!\mathrm{K}]^{+}$ and $[M]^{+}$ spectra, and its encoder rejects $10$ more. NPLIB1 does not record collision energy, so $35$\,eV is supplied for every spectrum. FIDDLE's released code trains its spectrum network but not its rescoring model, so FIDDLE runs as the network's prediction followed by its formula refinement, with the rescoring model replaced by a constant score that keeps the refinement order, which is the order of mass error.

%% =====================================================================
\section{Accuracy and Recall in Detail}\label{app:S8}

Table~\ref{tab:s-topk} extends Table~2 of the main text to top-10 accuracy. Table~\ref{tab:s-massbin} gives FACET's accuracy and recall by precursor mass in both adduct settings. Table~\ref{tab:s-sub-top1} breaks the top-1 accuracy down by adduct type, instrument type and composition for every method, and its last two columns give FACET's paired top-1 margin over MIST-CF and the exact McNemar $p$-value.

\begin{table}[htbp]
\caption{Top-$k$ accuracy and candidate-generation recall on the NPLIB1 test split ($n=2{,}194$).}
\label{tab:s-topk}
\centering
\begin{tabular}{llcccccc}
\toprule
Adduct & Method & Top-1 & Top-2 & Top-3 & Top-5 & Top-10 & Recall \\
\midrule
known & SIRIUS & 0.599 & 0.721 & 0.780 & 0.833 & 0.873 & 0.912 \\
 & FIDDLE & 0.641 & 0.667 & 0.675 & 0.678 & 0.678 & 0.678 \\
 & BUDDY & 0.736 & 0.767 & 0.776 & 0.783 & 0.788 & 0.789 \\
 & MIST-CF & 0.766 & 0.847 & 0.879 & 0.907 & 0.932 & 0.959 \\
 & FACET & 0.859 & 0.896 & 0.914 & 0.931 & 0.946 & 0.982 \\
\midrule
unknown & SIRIUS & 0.448 & 0.540 & 0.574 & 0.655 & 0.750 & 0.905 \\
 & BUDDY & 0.299 & 0.524 & 0.677 & 0.726 & 0.765 & 0.789 \\
 & MIST-CF & 0.660 & 0.779 & 0.838 & 0.887 & 0.927 & 0.951 \\
 & FACET & 0.729 & 0.867 & 0.893 & 0.912 & 0.936 & 0.994 \\
\bottomrule
\end{tabular}
\end{table}

\begin{table}[htbp]
\caption{FACET's top-$k$ accuracy and recall by precursor $m/z$, in both adduct settings.}
\label{tab:s-massbin}
\centering
\small
\begin{tabular}{lrcccccccc}
\toprule
 & & \multicolumn{4}{c}{Adduct known} & \multicolumn{4}{c}{Adduct unknown} \\
\cmidrule(lr){3-6}\cmidrule(lr){7-10}
Precursor $m/z$ & $n$ & Top-1 & Top-3 & Top-10 & Recall & Top-1 & Top-3 & Top-10 & Recall \\
\midrule
$<300$ & 536 & 0.924 & 0.961 & 0.987 & 0.996 & 0.841 & 0.951 & 0.983 & 0.996 \\
300--600 & 1{,}396 & 0.860 & 0.919 & 0.954 & 0.991 & 0.704 & 0.898 & 0.942 & 0.998 \\
600--800 & 198 & 0.768 & 0.854 & 0.864 & 0.909 & 0.687 & 0.798 & 0.864 & 0.995 \\
$>800$ & 64 & 0.562 & 0.609 & 0.688 & 0.891 & 0.469 & 0.594 & 0.641 & 0.891 \\
\bottomrule
\end{tabular}
\end{table}

\begin{table}[htbp]
\caption{Top-1 accuracy on subsets of the NPLIB1 test split, adduct known, with FACET's paired margin over MIST-CF and the McNemar $p$-value.}
\label{tab:s-sub-top1}
\centering
\footnotesize
\setlength{\tabcolsep}{3.5pt}
\begin{tabular}{lrcccccrr}
\toprule
Subset & $n$ & SIRIUS & FIDDLE & BUDDY & MIST-CF & FACET & $\Delta$ & $p$ \\
\midrule
all spectra & 2{,}194 & 0.599 & 0.641 & 0.736 & 0.766 & 0.859 & $+0.093$ & $<10^{-22}$ \\
\addlinespace
$[M\!+\!\mathrm{H}]^{+}$ & 1{,}650 & 0.630 & 0.614 & 0.702 & 0.751 & 0.833 & $+0.082$ & $<10^{-13}$ \\
$[M\!+\!\mathrm{Na}]^{+}$ & 185 & 0.427 & 0.670 & 0.751 & 0.768 & 0.897 & $+0.130$ & $<10^{-3}$ \\
$[M\!+\!\mathrm{NH_4}]^{+}$ & 150 & 0.393 & 0.873 & 0.847 & 0.833 & 0.973 & $+0.140$ & $<10^{-4}$ \\
$[M\!-\!\mathrm{H_2O}\!+\!\mathrm{H}]^{+}$ & 167 & 0.695 & 0.760 & 0.940 & 0.892 & 0.958 & $+0.066$ & 0.003 \\
$[M\!-\!2\mathrm{H_2O}\!+\!\mathrm{H}]^{+}$ & 35 & 0.571 & 0.314 & 0.914 & 0.657 & 0.943 & $+0.286$ & 0.002 \\
\addlinespace
Bruker Q-ToF & 714 & 0.633 & 0.856 & 0.838 & 0.882 & 0.966 & $+0.084$ & $<10^{-12}$ \\
Q-ToF, other & 407 & 0.553 & 0.528 & 0.828 & 0.715 & 0.853 & $+0.138$ & $<10^{-8}$ \\
Orbitrap & 334 & 0.656 & 0.464 & 0.623 & 0.683 & 0.805 & $+0.123$ & $<10^{-4}$ \\
not annotated & 718 & 0.564 & 0.572 & 0.630 & 0.719 & 0.781 & $+0.063$ & $<10^{-3}$ \\
\addlinespace
containing sulfur & 250 & 0.500 & 0.004 & 0.652 & 0.416 & 0.704 & $+0.288$ & $<10^{-14}$ \\
\bottomrule
\end{tabular}
\end{table}

Recall shows FACET's advantage where candidate generation matters most. For the $221$ spectra with a halogenated formula, FACET's recall is $1.000$ and that of MIST-CF $0.606$. For the $87$ spectra whose formula has two or more atoms of one halogen, which MIST-CF's decomposer excludes, the two recalls are $1.000$ and $0.000$.

%% =====================================================================
\section{The Errors That Remain}\label{app:S9}

\paragraph{Per-element count error.} Table~\ref{tab:s-elemerr} reports, for each element, the mean signed and mean absolute difference between the counts of the top-1 and the correct formula. A systematic bias would show as a large signed error. The largest is that of hydrogen, which is decoded last and closes the mass, so every exchange that involves hydrogen, such as the most frequent one, \ce{C3N} for \ce{O3H2}, shows in its count.

\begin{table}[htbp]
\caption{Mean signed and mean absolute error in the atom counts of the top-1 formula, per element ($n=2{,}194$).}
\label{tab:s-elemerr}
\centering
\begin{tabular}{lcccc}
\toprule
 & \multicolumn{2}{c}{Adduct known} & \multicolumn{2}{c}{Adduct unknown} \\
\cmidrule(lr){2-3}\cmidrule(lr){4-5}
Element & Signed & Absolute & Signed & Absolute \\
\midrule
\ce{C} & $+0.022$ & 0.351 & $-0.094$ & 0.373 \\
\ce{N} & $-0.007$ & 0.227 & $-0.037$ & 0.217 \\
\ce{O} & $+0.057$ & 0.282 & $+0.086$ & 0.414 \\
\ce{P} & $-0.015$ & 0.021 & $-0.015$ & 0.021 \\
\ce{S} & $-0.013$ & 0.057 & $+0.000$ & 0.065 \\
\ce{Cl} & $+0.018$ & 0.061 & $+0.005$ & 0.055 \\
\ce{F} & $+0.008$ & 0.072 & $+0.020$ & 0.089 \\
\ce{Br} & $-0.001$ & 0.004 & $+0.001$ & 0.007 \\
\ce{I} & $-0.006$ & 0.008 & $-0.005$ & 0.008 \\
\ce{H} & $-0.147$ & 0.408 & $-0.293$ & 0.717 \\
\bottomrule
\end{tabular}
\end{table}

\paragraph{Near-isobaric exchanges.} A top-1 error is mass-indistinguishable when the predicted and the correct formula both lie inside one mass window, so that mass could not have separated them. With the adduct known, FACET ranks $310$ of the $2{,}194$ spectra wrongly, and $276$ of these errors, or $89\%$, are mass-indistinguishable. With the adduct unknown, $61\%$ of its $594$ errors assign the wrong adduct, and $34\%$ are near-isobaric confusions under the correct adduct. Table~\ref{tab:s-swaps} lists the element-count differences behind the mass-indistinguishable errors. They are rarely the textbook single exchanges: the most frequent are compound exchanges with mass gaps of a few millidaltons or less, well inside the $30$\,ppm window.

\begin{table}[htbp]
\caption{The most frequent element-count differences between the top-1 and the correct formula among the mass-indistinguishable errors. ``Gap'' is the mass difference of the exchange.}
\label{tab:s-swaps}
\centering
\begin{tabular}{llccc}
\toprule
Setting & Exchange (top-1 $\leftrightarrow$ correct) & Gap (mDa) & Count & Share of errors (\%) \\
\midrule
adduct known & \ce{C3N} $\leftrightarrow$ \ce{O3H2} & 2.68 & 28 & 10.1 \\
adduct known & \ce{C2OH2} $\leftrightarrow$ \ce{N3} & 1.34 & 26 & 9.4 \\
adduct known & \ce{CN4} $\leftrightarrow$ \ce{O4H4} & 1.34 & 13 & 4.7 \\
adduct known & \ce{C2ClH} $\leftrightarrow$ \ce{N2S} & 1.54 & 12 & 4.3 \\
adduct known & \ce{C3O} $\leftrightarrow$ \ce{NClH3} & 0.49 & 11 & 4.0 \\
adduct known & \ce{C3} $\leftrightarrow$ \ce{OFH} & 1.14 & 9 & 3.3 \\
\midrule
adduct unknown & \ce{C3N} $\leftrightarrow$ \ce{O3H2} & 2.68 & 21 & 10.3 \\
adduct unknown & \ce{C2OH2} $\leftrightarrow$ \ce{N3} & 1.34 & 16 & 7.9 \\
adduct unknown & \ce{C3O} $\leftrightarrow$ \ce{NClH3} & 0.49 & 14 & 6.9 \\
adduct unknown & \ce{CN4} $\leftrightarrow$ \ce{O4H4} & 1.34 & 10 & 4.9 \\
adduct unknown & \ce{CSH2} $\leftrightarrow$ \ce{NO2} & 5.18 & 8 & 3.9 \\
adduct unknown & \ce{C2ClH} $\leftrightarrow$ \ce{N2S} & 1.54 & 8 & 3.9 \\
\bottomrule
\end{tabular}
\end{table}

\paragraph{Validity and adduct accuracy.} Validity asks whether a returned formula lies in the feasible set of the adduct the model used, which the automaton guarantees: FACET's validity rate is $1.000$ in both adduct settings. Adduct accuracy is a different quantity. With the adduct unknown, FACET's adduct accuracy is $0.835$, and checking the top-1 formula against the annotated adduct instead of the inferred one would measure this adduct accuracy, not validity.

%% =====================================================================
\section{Precursor Provenance}\label{app:S10}

NPLIB1 records the precursor of $1{,}410$ evaluated test spectra as the theoretical $m/z$ of the annotated ion and of $750$ as a measured value, which lies within $30$\,ppm of the theoretical one. For $34$ spectra the recorded precursor lies more than $30$\,ppm from the annotated formula and adduct. FACET and FIDDLE read the measured value where one is recorded, SIRIUS and BUDDY read the parent mass of the spectrum file, and MIST-CF reads the precursor of its published benchmark, the theoretical $m/z$ with simulated noise. On the measured-precursor spectra, the value FACET reads lies a median of $0.9$\,ppm from the theoretical $m/z$, and the value MIST-CF reads lies $1.5$\,ppm from it.

The provenance matters for ranking. Ranking each enumerated feasible set by mass error alone puts the correct formula first for $1{,}397$ of the $1{,}403$ theoretical-precursor spectra whose correct formula is enumerated, but for only $127$ of the $748$ measured-precursor spectra, so the measured-precursor spectra are where ranking quality is tested on equal terms. On them FACET leads MIST-CF in top-1 accuracy by $0.048$ with the adduct known and by $0.035$ with it unknown, with McNemar $p$-values of $0.0055$ and $0.062$. Table~\ref{tab:s-prec} gives accuracy and recall for every method by precursor type.

\begin{table}[htbp]
\caption{Top-1 accuracy and candidate-generation recall by how the precursor is recorded.}
\label{tab:s-prec}
\centering
\small
\begin{tabular}{llcccccc}
\toprule
 & & \multicolumn{2}{c}{Theoretical} & \multicolumn{2}{c}{Measured} & \multicolumn{2}{c}{Inconsistent} \\
\cmidrule(lr){3-4}\cmidrule(lr){5-6}\cmidrule(lr){7-8}
Adduct & Method & Top-1 & Recall & Top-1 & Recall & Top-1 & Recall \\
\midrule
known & SIRIUS & 0.635 & 0.931 & 0.556 & 0.913 & 0.029 & 0.088 \\
 & FIDDLE & 0.716 & 0.723 & 0.529 & 0.623 & 0.000 & 0.000 \\
 & BUDDY & 0.841 & 0.879 & 0.569 & 0.651 & 0.059 & 0.059 \\
 & MIST-CF & 0.793 & 0.966 & 0.723 & 0.947 & 0.618 & 0.912 \\
 & FACET & 0.926 & 0.996 & 0.771 & 1.000 & 0.000 & 0.000 \\
\midrule
unknown & SIRIUS & 0.428 & 0.923 & 0.503 & 0.909 & 0.029 & 0.088 \\
 & BUDDY & 0.375 & 0.879 & 0.168 & 0.651 & 0.029 & 0.059 \\
 & MIST-CF & 0.638 & 0.960 & 0.711 & 0.943 & 0.500 & 0.765 \\
 & FACET & 0.738 & 0.996 & 0.745 & 1.000 & 0.000 & 0.794 \\
\bottomrule
\end{tabular}
\end{table}

Of the $34$ spectra with an inconsistent precursor, $29$ are permanently charged cations, $26$ of them phospholipids, which are annotated as $[M\!+\!\mathrm{H}]^{+}$ although their precursor is the mass of the $[M]^{+}$ cation. The other $5$ lie $99$ to $10{,}071$\,ppm from every allowed adduct. MIST-CF recovers most of them because its benchmark derives the precursor from the annotation. With the adduct unknown, FACET also enumerates under $[M]^{+}$ and places $27$ of the $34$ correct formulas among its candidates.

%% =====================================================================
\section{Truncation, Decomposition Cost and Latency}\label{app:S11}

\begin{figure}[htbp]
\centering
\includegraphics[width=0.6\textwidth]{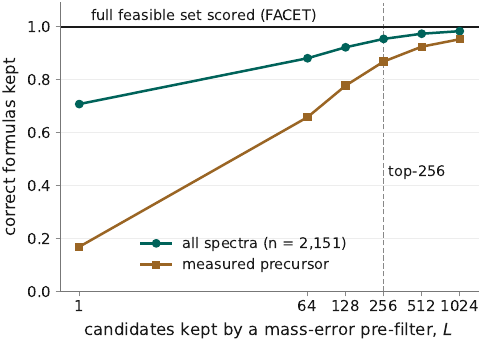}
\caption{Share of enumerated correct formulas kept by a cut to the $L$ feasible formulas closest in mass, for all test spectra and for those with a measured precursor.}
\label{fig:s-trunc}
\end{figure}

\paragraph{The cost of truncation.} Two-stage pipelines shorten the candidate list before their expensive scorer runs. To measure what this costs independently of any particular filter, we rank each enumerated feasible set by mass error alone and ask how often the correct formula survives a cut to the $L$ formulas closest in mass. Figure~\ref{fig:s-trunc} and Table~\ref{tab:s-trunc} give the answer for the $2{,}151$ test spectra whose correct formula is enumerated, and separately for the $748$ with a measured precursor. A cut to $256$ candidates loses $4.6\%$ of the correct formulas, and $13.1\%$ of those with a measured precursor. Scoring the full feasible set, as FACET does, keeps all of them.

\begin{table}[htbp]
\caption{Share of enumerated correct formulas kept by a mass-error pre-filter of size $L$.}
\label{tab:s-trunc}
\centering
\begin{tabular}{rcc}
\toprule
$L$ & All spectra & Measured precursor \\
\midrule
1 & 0.709 & 0.170 \\
64 & 0.881 & 0.659 \\
128 & 0.923 & 0.778 \\
256 & 0.954 & 0.869 \\
512 & 0.974 & 0.925 \\
1{,}024 & 0.984 & 0.953 \\
\midrule
full set & 1.000 & 1.000 \\
\bottomrule
\end{tabular}
\end{table}

\paragraph{Decomposition benchmark.} Table~4 of the main text compares the decomposers of FACET and SIRIUS~$6.5.4$ on $120$ neutral target masses from the test split, with a $30$\,ppm window and the same element bounds for both, namely \ce{C}~$0$--$74$, \ce{N}~$0$--$15$, \ce{O}~$0$--$33$, \ce{P}~$0$--$6$, \ce{S}~$0$--$9$, \ce{Cl}~$0$--$7$, \ce{F}~$0$--$21$, \ce{Br}~$0$--$4$, \ce{I}~$0$--$6$, \ce{H}~$0$--$128$. Both ran on one otherwise idle compute node. FACET on one thread returned $8{,}174{,}842$ formulas in $51.3$\,s of wall-clock time and $51.1$\,s of CPU time, or $159{,}863$ formulas per CPU second. With sixteen threads it took $15.8$\,s and $68.0$\,s of CPU time, used $4.3$ cores on average and produced $120{,}160$ formulas per CPU second. SIRIUS, given sixteen cores in one batch call, returned $9{,}060{,}070$ formulas in $47.7$\,s and $72.7$\,s of CPU time. It used $1.5$ cores on average and produced $124{,}574$ formulas per CPU second. On the $10$ heaviest masses alone, its wall-clock time did not respond to its core allocation, taking 40.3\,s on 3, 40.9\,s on 8 and 40.0\,s on 16 cores. Across $4$ repetitions of the benchmark, FACET's single-thread wall-clock time ranged over 50.4--51.3\,s, that of SIRIUS over 47.0--48.7\,s, and SIRIUS's throughput over 113,000--125,000 formulas per CPU second. SIRIUS returns $11\%$ more formulas for the same window, mostly because its RDBE filter admits $-0.5$, and in part because one of the masses exceeds FACET's enumeration budget, so that FACET returns its formulas only up to the budget. In the decoding pipeline such a mass is routed to beam search instead.

\paragraph{Latency.} Decoding latency was measured per spectrum on $2{,}176$ test spectra, all of them decoded exhaustively. The encoder pass takes a median of $4.6$\,ms, independent of the number of candidates. Spectra with at most $256$ candidates take a median of $7.9$\,ms in total. Beyond that size, the binned median latency grows in proportion to $|\Feas_a|^{0.86}$. Spectra with about $10^5$ candidates take a median of $3.4$\,s, and the largest feasible set of the split, $2.0$ million formulas, takes $17.6$\,s.

%% =====================================================================
\section{Representative Spectra}\label{app:S12}

Table~\ref{tab:s-cases} shows three test spectra, each chosen by a fixed rule to illustrate one aspect of FACET. Case (a) has a correct formula with two or more atoms of one halogen, which MIST-CF's decomposer cannot produce and FACET ranks first. Case (b) has a measured precursor and a correct formula that is not among the $256$ feasible formulas closest in mass, yet is ranked first by FACET. Case (c) has a measured precursor and a runner-up, a near-isobar within $3$\,mDa of the correct formula, that lies closer to the precursor in mass, and FACET still ranks the correct formula first. Each rule picks the most pronounced instance in the test split: for (a) the largest feasible set, for (b) the worst mass-error rank of the correct formula among spectra with feasible sets of up to $400{,}000$ formulas, and for (c) the largest mass-error advantage of the runner-up.

\begin{table}[htbp]
\caption{Three representative test spectra, adduct known. $|\Feas_a|$ is the size of the feasible set and ``Rank'' the rank FACET gives the correct formula. FACET's three highest-ranked formulas are given with their mass errors in ppm.}
\label{tab:s-cases}
\centering
\footnotesize
\setlength{\tabcolsep}{3pt}
\begin{tabular}{l>{\raggedright\arraybackslash}p{2.0cm}lcrc>{\raggedright\arraybackslash}p{4.5cm}>{\raggedright\arraybackslash}p{2.1cm}}
\toprule
 & Correct formula & Adduct & $m/z$ & $|\Feas_a|$ & Rank & FACET top three & MIST-CF top-1 \\
\midrule
(a) & \ce{C27H30Cl2O6} & $[M\!+\!\mathrm{H}]^{+}$ & 521.1490 & 12{,}625 & 1 & \ce{C27H30Cl2O6} ($0.4$), \ce{C24H31Cl2FO7} ($2.6$), \ce{C26H30ClO7P} ($0.1$) & \ce{C28H24O10} (correct formula absent) \\
(b) & \ce{C36H44O11} & $[M\!+\!\mathrm{Na}]^{+}$ & 675.2810 & 15{,}587 & 1 & \ce{C36H44O11} ($5.2$), \ce{C33H45FO12} ($3.5$), \ce{C33H48O11S} ($0.1$) & \ce{C36H44O11} \\
(c) & \ce{C14H22N2O3} & $[M\!+\!\mathrm{H}]^{+}$ & 267.1730 & 84 & 1 & \ce{C14H22N2O3} ($10.1$), \ce{C17H20N3} ($0.0$), \ce{C12H20N5O2} ($15.1$) & \ce{C14H22N2O3} \\
\bottomrule
\end{tabular}
\end{table}

In case (a), mometasone furoate (\ce{C27H30Cl2O6}) carries two chlorine atoms. Its feasible set holds $12{,}625$ formulas, and FACET ranks the correct formula first. MIST-CF's decomposer admits at most one atom of each halogen, so the correct formula never enters its candidate list, and it returns \ce{C28H24O10} instead.

In case (b), \ce{C36H44O11} is recorded with a measured precursor $5.2$\,ppm from its theoretical $m/z$. Its feasible set holds $15{,}587$ formulas, and by mass error alone the correct formula ranks only $2{,}715$th, so a pre-filter keeping the $256$ formulas closest in mass would have discarded it. FACET scores the entire set and ranks it first.

In case (c), practolol (\ce{C14H22N2O3}) is recorded with a precursor $10.1$\,ppm from its theoretical $m/z$, while the runner-up, \ce{C17H20N3}, matches the precursor to within $0.003$\,ppm. The two differ by $2.68$\,mDa, an exchange of \ce{C3N} for \ce{O3H2}. Mass error alone would rank the runner-up first, but the spectrum leads FACET to the correct formula.

\end{document}